\documentclass[12pt]{article}

\usepackage{amssymb,amsfonts,amsmath}
\usepackage{amsthm}
\usepackage{mathrsfs}

\usepackage{subcaption}
\usepackage{graphicx}
\usepackage{multirow}

\usepackage{overpic}
\usepackage{upgreek}

\usepackage{geometry}
\usepackage{xcolor}
\usepackage{multirow}
\usepackage{palatino}
\usepackage{tikz}
\usetikzlibrary{arrows,shapes,chains} 
\usepackage{authblk}
\usepackage{natbib}
\usepackage{setspace}
\usepackage{url}
\usepackage{hyperref}
\usepackage{titlesec}

\usepackage{algorithm}  
\usepackage{algorithmicx}
\usepackage[noend]{algpseudocode}

\usepackage{cases}

\newtheorem{theorem}{Theorem}
\newtheorem{definition}{Definition}

\newtheorem{lemma}{Lemma}

\titleformat{\title}{\LARGE\bfseries}{\thesection}{1em}{} 
\titleformat{\section}{\Large\bfseries}{\thesection}{1em}{}  
\titleformat{\subsection}{\normalsize\bfseries}{\thesubsection}{1em}{}  
\titleformat{\subsubsection}{\small\bfseries}{\thesubsubsection}{1em}{} 

\title{Efficiently Solving Discounted MDPs with Predictions on Transition Matrices}

\author{\small Lixing Lyu$^1$, Jiashuo Jiang$^2$, Wang Chi Cheung$^3$} 
\affil{\footnotesize 
$^1$Institute of Operations Research and Analytics, National University of Singapore \\
$^2$Department of Industrial Engineering and Decision Analytics, Hong Kong University of Science and Technology \\
$^3$Department of Industrial Systems Engineering and Management, National University of Singapore \footnote{Corresponding to Jiashuo Jiang, \textit{jsjiang@ust.hk}
and Wang Chi Cheung \textit{isecwc@nus.edu.sg}}
}

\allowdisplaybreaks[3]

\begin{document}

\bibliographystyle{plainnat}
\date{}  
\maketitle
\newcommand*\abs[1]{\lvert#1\rvert}

\begin{abstract}
We study infinite-horizon Discounted Markov Decision Processes (DMDPs)  under a generative model. Motivated by the Algorithm with Advice framework \cite{mitzenmacher2022algorithms}, we propose a novel framework to investigate how a prediction on the transition matrix can enhance the sample efficiency in solving DMDPs and improve sample complexity bounds. We focus on the DMDPs with $N$ state-action pairs and discounted factor $\gamma$. Firstly, we provide an impossibility result that, without prior knowledge of the prediction accuracy, no sampling policy can compute an $\epsilon$-optimal policy with a sample complexity bound better than $\tilde{O}((1-\gamma)^{-3} N\epsilon^{-2})$, which matches the state-of-the-art minimax sample complexity bound with no prediction. In complement, we propose an algorithm based on minimax optimization techniques that leverages the prediction on the transition matrix. Our algorithm achieves a sample complexity bound depending on the prediction error, and the bound is uniformly better than $\tilde{O}((1-\gamma)^{-4} N \epsilon^{-2})$, the previous best result derived from convex optimization methods. These theoretical findings are further supported by our numerical experiments.
\end{abstract}

\section{Introduction}

Markov decision process (MDP) is a fundamental mathematical framework for sequential decision making under uncertainty and serves as a essential tool for stochastic control and reinforcement learning (\cite{puterman2014markov,bertsekas2012dynamic,sutton2018reinforcement}). One of the most important problems in this field is to learn an approximately optimal policy for an MDP with high probability, given access only to a generative model. Specifically, this involves identifying a policy with a small error from the optimal policy in terms of expected cumulative reward over an infinite horizon through sampling state transitions. Two main classes have been extensively studied: discounted MDP (DMDP) \cite{wang2020randomized,jin2020efficiently,sidford2018near,wainwright2019variance} and average-reward MDP (AMDP)\cite{wang2017primal,jin2020efficiently,jin2021towards}. 

In this paper, we focus on this specific problem of infinite horizon DMDPs. A DMDP instance is desribed as a tuple $\mathcal{M} = (\mathcal{S},\mathcal{A},\boldsymbol{\text{P}},\boldsymbol{\text{r}},\gamma)$. Specifically, $\mathcal{S}$ is a known finite state space with size $|\mathcal{S}|$. The collection $\mathcal{A} = \{ \mathcal{A}_i\}_{i \in \mathcal{S}}$ is a known collection of finite set of actions $\mathcal{A}_i$ for state $i \in \mathcal{S}$. For simplicity, we denote $\mathcal{N} = \{(i,a):i \in \mathcal{S},a \in \mathcal{A}_i\}$ be the set of all state-action pairs, with size $|\mathcal{N}|=N$. The matrix $\boldsymbol{\text{P}} = (p(j|i,a))_{(i,a) \in \mathcal{N},j \in \mathcal{S}}  \in \mathbb{R}^{N  \times|\mathcal{S}|}$ is the unknown state-action-state matrix.
The $((i,a),j)$-th indice $p(j|i,a) \in [0,1]$ represents the transition probability from state $i$ to state $j$ under action $a$. 
For any state-action pair $(i,a) \in \mathcal{N}$, $\sum_{j \in \mathcal{S}} p(j|i,a) = 1$.
The array $\boldsymbol{\text{r}} = (\text{r}_{i,a})_{i \in \mathcal{S},a \in \mathcal{A}}$ is the known vector of state-action reward, where $\text{r}_{i,a} \in [0,1]$ be the reward received when taking action $a$ at state $i$. 
Finally, $\gamma \in (0,1)$ is the known discounted factor of the MDP. 
Following the literature, we assume that we can access to a generative model, allowing us to sample state transitions of one DMDP model. 
\textcolor{black}{In this setting, the sample complexity refers to the total number of samples required to compute an optimal policy that maximizes the expected cumulative reward with an error of at most $\epsilon$. The objective is to minimize this sample complexity.}



This problem has been extensively studied in the literature. \cite{gheshlaghi2013minimax} establish a sample complexity lower bound of $\tilde{\Omega}((1-\gamma)^{-3} N \epsilon^{-2})$ for finding an $\epsilon$-optimal policy. This bound is achieved by methods such as variance-reduced value iteration \cite{sidford2018near} and Q-learning \cite{wainwright2019variance}. Additionally, \cite{jin2020efficiently} proposes a stochastic mirror descent algorithm for both DMDP and AMDP and achieves a sample complexity bound of $\tilde{O}((1-\gamma)^{-4} N \epsilon^{-2})$ for DMDP, which currently represents the best achievable sample complexity bound by primal-dual type methods. 

Motivated by the Online Algorithm with Advice framework \cite{mitzenmacher2022algorithms} and Reinforcement Learning with Advice \cite{golowich2022can}, 
we consider the problem of learning infinite horizon DMDPs with possibly biased predictions. The process consists of two phases: a preparation phase and a learning phase. In the preparation phase, we receive a \textit{prediction} matrix $\hat{\boldsymbol{\text{P}}}$ for latent transition matrix $\boldsymbol{\text{P}}$. The subsequent learning phase mirrors the standard problem of learning infinite horizon DMDPs with a generative model, with the added feature that the prediction matrix $\hat{\boldsymbol{\text{P}}}$ can be incorporated in decision-making. The objective remains consistent with the standard problem: maximizing the expected cumulative discounted rewards over an infinite horizon.

Intuitively, when the prediction matrix $\hat{\boldsymbol{\text{P}}}$ is significantly different from the true $\boldsymbol{\text{P}}$, the learning algorithms should ignore the prediction entirely. Conversely, when $\hat{\boldsymbol{\text{P}}}$ is sufficiently close to $\boldsymbol{\text{P}}$, the learning algorithms can leverage the information from $\hat{\boldsymbol{\text{P}}}$ to reduce unnecessary sampling. For example, in the extreme case where $\hat{\boldsymbol{\text{P}}} = \boldsymbol{\text{P}}$, the optimal policy can be directly obtained through value iteration, policy iteration or linear programming \cite{puterman2014markov}, eliminating the need for any additional sampling.

The above discussion raises the following question (Que): 
\textit{Can we design an algorithm that efficiently leverages the prediction to improve the sample complexity bound, while remaining robust to cases where the prediction is absent or inaccurate?}

However, the answer to this question is mixed. To address it, we make the following novel contributions:

\textbf{Impossibility Result} In Section \ref{sec:impossibility-result}, we demonstrate that, even with predictions, no algorithm can compute an $\epsilon$-optimal policy with fewer than or equal to $\tilde{o}((1-\gamma)^{-3} N \epsilon^{-2})$ samples for all DMDP instances without prior knowledge of the prediction error of $\hat{\boldsymbol{\text{P}}}$. This result implies that no algorithm utilizing predictions can outperform the approaches proposed by \cite{sidford2018near,wainwright2019variance,agarwal2020model}, which achieve a sample complexity bound of $\tilde{O}((1-\gamma)^{-3} N \epsilon^{-2})$. Consequently, the answer to the question (Que) we raise above is ``NO".

\textbf{Levarage Prediction without knowing the prediction error} In Section \ref{sec:algorithm-noupper}, we take a small step back and focus on primal-dual type algorithms.
We propose \underline{Op}timistic-\underline{P}redict \underline{M}irror \underline{D}escent (OpPMD), for computing an $\epsilon$-optimal policy and utilizing prediction transition matrix $\hat{\boldsymbol{\text{P}}}$ judiciously. OpPMD novelly incorporates the prediction into estimation for future gradients, using carefully designed learning rates that eliminate the need for prior knowledge of the prediction error and the desired accuracy $\epsilon$. We derive a sample complexity bound for OpPMD that depends on the prediction error, despite the algorithm not knowing on the error. This bound is uniformly better than $\tilde{O}((1-\gamma)^{-4} N \epsilon^{-2})$, the current state-of-the-art bound achieved by primal-dual methods. 

\textbf{Numerical Validations} In Section \ref{sec:numerical}, we perform numerical experiments on a simple MDP instance to validate our approach. The result demonstrates the benefits of predictions and illustrates how effectively our algorithm leverages the prediction judiciously.

Finally, we provide discussions on future directions in Appendix \ref{sec:app-discussions-future}.


\subsection{More about Prediction $\hat{\boldsymbol{\text{P}}}$}

In this paper, we primarliy investigate whether predictions on transition matrix $\boldsymbol{\text{P}}$ can enhance the process of computing an (approximately) optimal policy and improve sample complexity bound. The prediction $\hat{\boldsymbol{\text{P}}}$ is expressed as a matrix $\hat{\boldsymbol{\text{P}}} = (\hat{p}(j|i,a))_{(i, a)\in \mathcal{N}, j\in \mathcal{S}} \in \mathbb{R}^{|\mathcal{N}| \times|\mathcal{S}|}$, with the same dimensions as the true transition matrix $\boldsymbol{\text{P}}$. Here $\hat{p}(j|i,a) \in [0,1]$ for all $i,a,j$ and $\sum_{j \in \mathcal{S}} \hat{p}(j|i,a)=1$ for all $(i,a) \in \mathcal{N}$. The prediction transition matrix $\hat{\boldsymbol{\text{P}}}$ is provided during the preparation phase, prior to the standard sampling and learning phase. The prediction error $\text{Dist}(\cdot,\cdot)$ is defined as follows, which is not known in advance:
\begin{equation*}
    \text{Dist}(\boldsymbol{\text{P}},\hat{\boldsymbol{\text{P}}}) = \max_{(i,a) \in \mathcal{N}}  \sum_{j \in \mathcal{S}} \left | \hat{p}(j|i,a))  - p(j|i,a))  \right|.
\end{equation*}
The prediction is a general Black-Box model, and we do not impose any additional assumptions on the prediction $\hat{\boldsymbol{\text{P}}}$. In practice, this prediction can be provided by some external sources. There are several potential candidates for constructing such prediction $\hat{\boldsymbol{\text{P}}}$. For example, suppose we already have a set of transition observations $\{(i_{\ell},a_{\ell},j_{\ell})\}_{\ell \in [T_0]}$ from an MDP with a transition matrix that is possibly similar to the DMDP model we aim to learn. Then we can compute $\hat{\boldsymbol{\text{P}}}$ as the sample mean for each tuple of $(i,a,j)$ as
\begin{equation*}
    \hat{p}(j|i,a) = \frac{|\{\ell:(i_{\ell},a_{\ell},j_{\ell}) = (i,a,j)\}|}{\max\{|\{\ell:(i_{\ell},a_{\ell}) = (i,a)\}|,1\}}.
\end{equation*}
Several powerful physics simulation engines in robotics, such as MuJoCo \cite{todorov2012mujoco}, can be used to collect such datasets. Besides, in healthcare applications, data can be collected from previous iterations of clinical trials for treating specific diseases. Furthermore, the prediction is related to transfer learning and multi-task learning. We provide more discussions in Appendix \ref{sec:app-discuss-prediction-P}.

\subsection{Techniques Overview: How to Leverage $\hat{\boldsymbol{\text{P}}}$?}

Following the line of \cite{wang2017primal,wang2020randomized,jin2020efficiently}, we formulate the infinite horizon DMDP problem as a minimax bilinear optimization problem using linear duality. To leverage the prediction effectively, we introduce several novel improvements. 
Firstly, we construct predicted gradients based on the prediction $\hat{\boldsymbol{\text{P}}}$, using it as a proxy of the future gradients within an optimistic mirror descent framework \cite{rakhlin2013online,joulani2017modular,orabona2019modern}. 
\textcolor{black}{Together with the standard approach of constructing stochastic unbiased gradients from sampled state transitions, our method introduces a novel combination of these two gradient sources.} This approach ensures that we benefit from accurate predictions while maintaining the same convergence rate as prediction-free cases when the prediction is uninformative.
Second, by carefully tuning the learning rate, our algorithm becomes parameter-free, eliminating the dependence on the prediction error and the desired accuracy level in implementation and enhancing practical usability.
Third, we provide a simple and new variance reduction technique to control the variance of stochastic estimators for the gradients and shrink them as the sample size increases, ensuring efficiency. Putting all these together into the minimax optimization framework, we provide a robust primal-dual type algorithm, showing that it can leverage the prediction $\hat{\boldsymbol{\text{P}}}$ efficiently with a more favorable sample complexity bound.

\subsection{Related Work}


Dynamic Programming and Reinforcement Learning have been well studied for decades with numerous applications in industry and finance. For further details, see textbooks \cite{bellman1966dynamic,bertsekas1996neuro,bertsekas2012dynamic,bertsekas2022abstract,sutton2018reinforcement,szepesvari2022algorithms} and surveys \cite{kaelbling1996reinforcement,arulkumaran2017deep}.

Solving MDPs and finding optimal policies efficiently is one of the most classic problems in dynamic programming and reinforcement learning. 
Recent research focuses on this specific problem with access only to a generative model for sampling state transitions. Both DMDP and AMDP have been well-studied under this assumption. For DMDP, \cite{gheshlaghi2013minimax} provide a sample complexity lower bound of $\tilde{\Omega}((1-\gamma)^{-3} N \epsilon^{-2})$. There are several works that match this bound, including value iteration \cite{sidford2018near}, Q-learning \cite{wainwright2019variance}, and Empirical MDP with Blackbox \cite{agarwal2020model}.
However, no primal-dual algorithms have yet achieved the optimal bound. \cite{wang2020randomized} proposes a randomized primal-dual method with sample complexity $\tilde{\Omega}(\min\{(1-\gamma)^{-6} |\mathcal{S}|^2N \epsilon^{-2},\tau^{4}(1-\gamma)^{-4} N \epsilon^{-2}\})$, where $\tau$ is an ergodic condition parameter. 
\cite{jin2020efficiently,cheng2020reduction} achieve $\tilde{O}((1-\gamma)^{-4} N \epsilon^{-2})$ using mirror descent type algorithms, respectively, representing the best results for primal-dual methods. Our work is closely related to \cite{jin2020efficiently}, and we provide more discussions in Appendix \ref{sec:app-com-sidford-efficient}.
Whether convex optimization methods can achieve the optimal sample complexity remains an open problem.
Our work takes a different approach: We explore whether incorporating predictions of the transition matrix can enhance primal-dual methods and improve sample complexity bounds. To the best of our knowledge, this is the first work to introduce the notion of prediction into solving MDPs with access only to a generative model.



Our work closely relates to reinforcement learning with prediction. \cite{golowich2022can} consider tabular MDPs with advice in the form of prediction of the Q-value and shows that this information can reduce regret. \cite{li2024beyond} study single-trajectory time-varying MDPs with untrusted machine-learned prediction. \cite{cutkosky2022leveraging} leverages hint in the form of guessed optimal actions under a stochastic linear bandit setting. \cite{lyu2023bandits} explore how dynamic predictions can enhance algorithm performance regarding regret in a non-stationary bandit with knapsack problem. However, all these works use regret as the primary performance metric. In contrast, our problem is the first to focus on utilizing black-box predictions to reduce unnecessary exploration and improve sample complexity bound in solving MDPs and reinforcement learning. Our work aligns with the research stream of Algorithm with Advice \cite{mitzenmacher2022algorithms} in general, and we provide more discussions in Appendix \ref{sec:app-com-alg-advice}.





Our algorithmic design is inspired by the parameter-free methods and adaptive learning rates from deterministic optimization. \cite{carmon2022making} develop a parameter-free stochastic gradient descent algorithm that achieves a double-logarithmic factor overhead compared to the optimal rate in the known-parameter setting. \cite{streeter2010less} introduces adaptive learning rates for online gradient descent. Interested readers can consult \cite{auer2002adaptive,orabona2014simultaneous,cutkosky2018black,orabona2019modern} for more references.


Finally, our work is related to the line of Offline Reinforcement Learning \cite{levine2020offline}. In this setting, the learning algorithm cannot interact with the environment and collect additional information about the model. Instead, it is provided by a static dataset of transitions and must learn the best policy based on this dataset. Various approaches have been developed for this problem, such as \cite{fujimoto2019off,kumar2019stabilizing,kumar2020conservative,agarwal2020optimistic,wu2019behavior}. While such datasets could potentially be used to construct a prediction transition matrix (e.g., via sample mean), these methods cannot apply to our setting, as our model allows for interaction with the environment through sampling state transitions. Furthermore, these approaches typically leverage datasets in ways far from estimating the transition matrix, limiting their applicability to our problem, where no dataset is available in advance.



\subsection{Notations}

For a positive integer $n$, we denote $[n] = \{1,2,...,n\}$, and let $\Delta^n = \{\boldsymbol{x} \in \mathbb{R}^n: \boldsymbol{x} \ge \boldsymbol{0}, \sum_{i=1}^n x_i = 1\}$ represent the $n$-dimensional probabilistic simplex. We use the $O(\cdot)$, $o(\cdot)$, $\Omega(\cdot)$ notations as defined in \cite{cormen2022introduction}. We adopt $\tilde{O}(\cdot)$, $\tilde{o}(\cdot)$, $\tilde{\Omega}(\cdot)$ notations, which has the same meaning as $O(\cdot)$, $o(\cdot)$, $\Omega(\cdot)$, respectively, except that they hide the logarithmic terms.

\section{Preliminaries}

\subsection{Value function}


Consider a DMDP instance $\mathcal{M} = (\mathcal{S},\mathcal{A},\boldsymbol{\text{P}},\boldsymbol{\text{r}},\gamma)$. 
A stationary (and randomized) policy 
can be represented by a collection of probability distributions $\pi = \{\pi_i\}_{i \in \mathcal{S}}$, where $\pi_i \in \Delta^{|\mathcal{A}_i|}$ is a probability distribution over $\mathcal{A}_i$, and $\pi_i(a)$ refers to the probability of taking $a \in \mathcal{A}_i$ at state $i$. For a policy $\pi$, we define the value function vector $\boldsymbol{v}^{\pi} = (v_i^{\pi})_{i \in \mathcal{S}} \in \mathbb{R}^{|\mathcal{S}|}$ as
\begin{equation*}
    v_i^{\pi} =  \mathbb{E}^{\pi} \left[\sum_{t=0}^{\infty} \gamma^t \text{r}_{i_t,a_t} | i_0 = i  \right], \quad \forall i \in \mathcal{S}.
\end{equation*}
The expectation operator $\mathbb{E}^{\pi}$ is taken over the state-action trajectory  $(i_0,a_0,i_1,a_1,\cdots)$ generated by the MDP under policy $\pi$. The optimal value vector $\boldsymbol{v}^{*} = (v_i^{*})_{i \in \mathcal{S}} \in \mathbb{R}^{|\mathcal{S}|}$ is defined as $\forall i \in \mathcal{S}$
\begin{equation*}
    \begin{aligned}
     v_i^{*} & =  \max_{\pi} \ \mathbb{E}^{\pi} \left[\sum_{t=0}^{\infty} \gamma^t \text{r}_{i_t,a_t} | i_0 = i  \right] \\
     & = \mathbb{E}^{\pi^*} \left[\sum_{t=0}^{\infty} \gamma^t \text{r}_{i_t,a_t} | i_0 = i  \right].
     \end{aligned}
\end{equation*}
where $\pi^*$ is the optimal stationary policy that achieves $\boldsymbol{v}^{*}$. \textcolor{black}{Given an initial distribution $\boldsymbol{q} \in \Delta^{|\mathcal{S}|}$, the value function with respect to $\boldsymbol{q}$ for a policy $\pi$ is defined as
\begin{equation*}
    v^{\pi}(\boldsymbol{q}) = \mathbb{E}^{\pi} \left[\sum_{t=0}^{\infty} \gamma^t \text{r}_{i_t,a_t} | i_0 \sim \boldsymbol{q} \right] = \boldsymbol{q}^{\top} \boldsymbol{v}^{\pi}.
\end{equation*}
with its optimal $v^{*}(\boldsymbol{q}) = \boldsymbol{q}^{\top} \boldsymbol{v}^*$. A policy $\pi^{(\epsilon)}$ is $\epsilon$-optimal with respect to $\boldsymbol{q}$ if $v^{*}(\boldsymbol{q}) \le v^{\pi^{(\epsilon)}}(\boldsymbol{q}) + \epsilon$.}


\subsection{Bellman Equation, Linear Programming (LP) Formulation and Minimax Formulation}


According to \cite{bertsekas2012dynamic,puterman2014markov}, $\boldsymbol{v}^*$ is the optimal value vector of an DMDP if and only if it satisfies the following \textit{Bellman equations}

\begin{equation}
    v_i^* = \max_{a \in \mathcal{A}_i} \left \{\gamma \sum_{j \in \mathcal{S}} p(j|i,a) v_{j}^* + \text{r}_{i,a} \right \},\quad \forall i \in \mathcal{S}.
    \label{eq:bellman-dmdp}
\end{equation}

When $\gamma \in (0,1)$, the Bellman equation has a unique fixed point solution. A policy $\pi^*$ is an optimal policy for the DMDP if it achieves $\boldsymbol{v}^*$ coordinate-wise, i.e. $\boldsymbol{v}^* = \boldsymbol{v}^{\pi^*}$. For finite-state DMDP, such optimal policy $\pi^*$ always exists. \cite{puterman2014markov} shows that the Bellman equation (\ref{eq:bellman-dmdp}) is equivalent to the following Linear Programming (LP) problem:
\begin{equation}
    \label{eq:lp-primal-bellman-dmdp}
    \begin{aligned}
        \min_{\boldsymbol{v}\in \mathbb
        {R}^{|\mathcal{S}|}} \quad & (1 - \gamma) \boldsymbol{q}^{\top} \boldsymbol{v} \\
        \text{s.t.} \quad & (\hat{\boldsymbol{\text{I}}} - \gamma \boldsymbol{\text{P}}) \boldsymbol{v} - \boldsymbol{\text{r}} \ge \boldsymbol{0}.
    \end{aligned}
\end{equation}
where $\hat{\boldsymbol{\text{I}}} \in \mathbb{R}^{|\mathcal{N}| \times|\mathcal{S}|}$ be the matrix defined as $I_{(i,a),j} = \boldsymbol{1}[i = j]$ for each $((i,a),j)\in \mathcal{N} \times \mathcal{S}$. The dual problem of (\ref{eq:lp-primal-bellman-dmdp}) is 
\begin{equation}
    \label{eq:lp-dual-bellman-dmdp}
    \begin{aligned}
        \textcolor{black}{\max_{\boldsymbol{\mu}\in \Delta^N}} \quad &  \boldsymbol{\mu}^{\top} \boldsymbol{\text{r}} \\
        \text{s.t.} \quad &  (\hat{\boldsymbol{\text{I}}} - \gamma \boldsymbol{\text{P}})^{\top} \boldsymbol{\mu} = (1 - \gamma) \boldsymbol{q}.
    \end{aligned}
\end{equation}

Denote $\boldsymbol{\mu}^*$ as an optimal solution of (\ref{eq:lp-dual-bellman-dmdp}). We can formulate (\ref{eq:lp-primal-bellman-dmdp}), (\ref{eq:lp-dual-bellman-dmdp}) as the following equivalent minimax problem
\begin{equation}
    \label{eq:lp-minimax-bellman-dmdp}
    \begin{aligned}
        \min_{\boldsymbol{v} \in \mathcal{V}} &  \max_{\boldsymbol{\mu} \in \mathcal{U}} \quad  f(\boldsymbol{v},\boldsymbol{\mu}),\\
        \text{s.t.}& \quad  f(\boldsymbol{v},\boldsymbol{\mu}) = (1 - \gamma) \boldsymbol{q}^{\top} \boldsymbol{v} + \boldsymbol{\mu}^{\top} ((\gamma \boldsymbol{\text{P}} - \hat{\boldsymbol{\text{I}}}) \boldsymbol{v} + \boldsymbol{\text{r}}).
    \end{aligned}
\end{equation}
where $\mathcal{V} = \{\boldsymbol{v} \in \mathbb{R}^m: \|\boldsymbol{v}\|_{\infty} \le (1-\gamma)^{-1}\}$, $\mathcal{U} = \Delta^{N}$. It is straightforward to verify that $\boldsymbol{v}^* \in \mathcal{V}$ and $\boldsymbol{\mu}^* \in \mathcal{U}$. Since $\text{r}_{i,a} \in [0,1]$ for all $(i,a)$, it follows that $v_i^* \le (1-\gamma)^{-1}$ for all $i \in \mathcal{S}$. By multiplying $\boldsymbol{e}^{\top}$ (where $\boldsymbol{e} = (1,\cdots,1)$) on both sides of the constraint $(\hat{\boldsymbol{\text{I}}} - \gamma \boldsymbol{\text{P}})^{\top} \boldsymbol{\mu} = (1 - \gamma) \boldsymbol{q}$, we can confirm that $\boldsymbol{e}^{\top} \boldsymbol{\mu}^* = 1$, as $\boldsymbol{q}$ is a probability distribution.

\section{An Impossibility Result}
\label{sec:impossibility-result}

In this section, we demonstrate that even with a prediction, no algorithm can compute an $\epsilon$-optimal policy for all DMDP instances with strictly fewer than $\tilde{O}((1-\gamma)^{-3} N \epsilon^{-2})$ samples without additional knowledge of the prediction error, i.e., the discrepancy between $\hat{\boldsymbol{\text{P}}}$ and the true $\boldsymbol{\text{P}}$. The analysis in this section is inspired by a lower bound in the prediction-free setting derived by \cite{gheshlaghi2013minimax}, and the high-level ideas from the Pareto regret frontier for multi-armed bandits \cite{lattimore2015pareto}. We focus on the $(\epsilon,\delta)$-\textit{smart} algorithm \textcolor{black}{whose objective is to output a near-optimal policy with high probability through sampling}, as defined below:

\begin{definition}
    \textcolor{black}{
    We say an algorithm $\text{ALG}$ is a $(\epsilon,\delta)$-\textit{smart} algorithm with respect to $T$ on a DMDP model $\mathcal{M}$, if it outputs a policy $\hat{\pi}_T$ such that
    \begin{equation}
        \Pr_{\mathcal{M},T} \left( \left\| \boldsymbol{v}^{\hat{\pi}_T} - \boldsymbol{v}^* \right \|_{\infty}  >\epsilon \right) < \delta.
        \label{eq:epsilon-delta-smart-condition}
    \end{equation}
    Here $\hat{\pi}_T$ is generated by $\text{ALG}$ after adaptively sampling $T$ state-action pairs and observing the subsequent transitions on the model $\mathcal{M}$.}
\end{definition}

We remark that an $(\epsilon,\delta)$-\textit{smart} algorithm guarantees to find an $\epsilon$-optimal value function with high probability, rather than an $\epsilon$-optimal policy with high probability. 
In other words, the output policy $\hat{\pi}_T$ by an $(\epsilon,\delta)$-\textit{smart} algorithm only satisfies (\ref{eq:epsilon-delta-smart-condition}), but is not guaranteed to be $\epsilon$-optimal.
Notably, while an $\epsilon$-optimal value function does not necessarily imply an $\epsilon$-optimal policy, the converse is true. Therefore, a lower bound on the sample complexity for finding an $\epsilon$-optimal value function also serves as a lower bound for finding an $\epsilon$-optimal policy. 

Now, we are ready to state our impossibility result. 

\begin{theorem}
    Suppose $N \ge 6$, $\gamma \in \left[1/3,1\right)$, $\epsilon \in \left (0,(1-\gamma)^{-1}/40 \right]$, $\delta \in (0,0.24]$. 
    Consider a fixed but arbitrary algorithm ALG. If ALG is $(\epsilon,\delta)$-\textit{smart} on a specific DMDP instance $\mathcal{M}_0$ with $\mathcal{S}$, $\mathcal{A}$ with total $N$ state-action pairs, discounted factor $\gamma$, and transition matrix $\boldsymbol{\text{P}}_0$ given access to a black-box prediction $\hat{\boldsymbol{\text{P}}}$ satisfying $\hat{\boldsymbol{\text{P}}} = \boldsymbol{\text{P}}_0$, and the number of samples $T$ satisfies
    \begin{equation}
        T \le \frac{1}{300}(1-\gamma)^{-3} \left(\frac{N}{3}-1 \right) \epsilon^{-2} \ln\left(\frac{1}{4.1\delta}\right),
        \label{eq:impossibility-result-T}
    \end{equation}
    then there exists another DMDP instance $\mathcal{M}'$ \textcolor{black}{with the same ${\cal S}, {\cal A}, \gamma$ as $\mathcal{M}$} such that, even with the same prediction $\hat{\boldsymbol{\text{P}}}$, ALG fails to be $(\epsilon,\delta)-$\textit{smart} on $\mathcal{M}'$ with the same $T$.
    \label{thm:impossibility-result}
\end{theorem}

Theorem \ref{thm:impossibility-result} is proved in Appendix \ref{sec:app-pf-thm-impossibility}. 
\textcolor{black}{In fact, even if ALG knows that the instance is either $\mathcal{M}_0$ or $\mathcal{M}'$, which share the same prediction $\hat{\boldsymbol{\text{P}}}$, but does not know the exact identity of the instance, the impossibility result still holds. This shows the fundamental challenge of leveraging predictions.}

We provide some intuition for the proof of Theorem \ref{thm:impossibility-result}. First, \textcolor{black}{we construct a specific class $\mathcal{I}$ of DMDP instances, and this class $\mathcal{I}$ contains instance $\mathcal{M}_0$ with transition matrix $\boldsymbol{\text{P}}_0 = \hat{\boldsymbol{\text{P}}}$} 
(although the algorithm does not know that 
\textcolor{black}{$\boldsymbol{\text{P}}_0 = \hat{\boldsymbol{\text{P}}}$}). Next, we construct another DMDP instance $\mathcal{M}' \in \mathcal{I}$ with transition matrix $\boldsymbol{\text{P}}'$, such that (1) $\text{Dist}(\boldsymbol{\text{P}}_0,\boldsymbol{\text{P}}') \le O ((1-\gamma)^2 \epsilon)$, but (2) the optimal policies for $\mathcal{M}_0$, $\mathcal{M}'$ are fundamentally different. Using KL divergence techniques, we illustrate that if the sample number $T$ satisfies (\ref{eq:impossibility-result-T}), then no algorithm can simultaneously establish both that (1) $\hat{\boldsymbol{\text{P}}}$ is useful for learning $\mathcal{M}_0$, and (2) $\hat{\boldsymbol{\text{P}}}$ is misleading for learning $\mathcal{M}'$. This leads to the conclusion of Theorem \ref{thm:impossibility-result}. More detailed discussion is provided in Appendix \ref{sec:app-discuss-thm-impossibility}, together with the proof. The proof reveals the high-level idea behind Theorem \ref{thm:impossibility-result}: The cost of identifying whether the prediction $\hat{\boldsymbol{\text{P}}}$ matches the true $\boldsymbol{\text{P}}$ is relatively high compared to simply ignoring $\hat{\boldsymbol{\text{P}}}$ and sampling from scratch.



\section{Leverage Black-Box Prediction $\hat{\boldsymbol{\text{P}}}$}

\label{sec:algorithm-noupper}

In this section, we propose an algorithm, \underline{Op}timistic-\underline{P}redict \underline{M}irror \underline{D}escent (OpPMD), presented in Algorithm \ref{alg:minimax-mirror-noupper}, that leverages the prediction $\hat{\boldsymbol{\text{P}}}$ without requiring the knowledge of the prediction error. Our approach hinges on a carefully designed primal-dual mirror descent method for solving the minimax problem (\ref{eq:lp-minimax-bellman-dmdp}) while incorporating $\hat{\boldsymbol{\text{P}}}$. 
In Section \ref{sec:alg-noupper-gradient-estimators}, we provide the gradient estimators required for mirror descent on $\boldsymbol{v}$- and $\boldsymbol{\mu}$- sides, respectively. Section \ref{sec:alg-noupper-description} details our OpPMD. In Section \ref{sec:alg-noupper-analysis-sample-complexity}, we analyze our algorithm, providing a sample complexity bound that depends on the prediction error $\text{Dist}(\boldsymbol{\text{P}},\hat{\boldsymbol{\text{P}}})$, even though OpPMD does not know $\text{Dist}(\boldsymbol{\text{P}},\hat{\boldsymbol{\text{P}}})$. 
In Section \ref{sec:alg-noupper-sketch-proof-minimax}, we provide a sketch proof of the quality of the minimax solution output by our algorithm in terms of the duality gap. For simplicity, throughout the section we abbreviate $\text{Dist}(\boldsymbol{\text{P}},\hat{\boldsymbol{\text{P}}})$ as $\text{Dist}$.

\begin{algorithm}[htb]
	\caption{Optimistic-Predict-Mirror Descent (OpPMD)}
	\begin{algorithmic}[1]
	    \State \textbf{Input:} Optimization length $T$, initialized $\boldsymbol{v}_1 \in \mathcal{V}$, $\boldsymbol{\mu}_1 = (\frac{1}{N},\cdots \frac{1}{N}) \in \mathcal{U}$, initial distribution $\boldsymbol{q}  \in \Delta^{|\mathcal{S}|}$, prediction matrix $\hat{\boldsymbol{\text{P}}}$, $\bar{\boldsymbol{g}}_{1}^{\boldsymbol{\mu}} = (\hat{\boldsymbol{\text{I}}} - \gamma \hat{\boldsymbol{\text{P}}}) \boldsymbol{v}_{1} - \boldsymbol{\text{r}}$.
       \For{$t \in [T]$} 
       \State Sample and compute $\tilde{\boldsymbol{g}}^{\boldsymbol{v}}_t$ as (\ref{eq:smd-prediction-noupper-v-gradient}).
       \State Compute learning rate for $\boldsymbol{v}-$side:
       \begin{equation}
           \eta_t^v = \frac{\sqrt{2}}{2} \cdot \frac{\sqrt{|\mathcal{S}|} \cdot (1-\gamma)^{-1}}{\sqrt{\sum_{i=1}^t \|\tilde{\boldsymbol{g}}^{\boldsymbol{v}}_i\|_2^2}}.
           \label{eq:smd-prediction-noupper-v-learning-rate}
       \end{equation}
       \State Update $\boldsymbol{v}_{t+1}$: 
       \begin{equation}
           \boldsymbol{v}_{t+1}  = \Pi_{\mathcal{V}} (\boldsymbol{v_t} - \eta_t^v\tilde{\boldsymbol{g}}^{\boldsymbol{v}}_t).
           \label{eq:smd-prediction-noupper-v-update}
       \end{equation}
       \State Sample and compute $\tilde{\boldsymbol{g}}^{\boldsymbol{\mu}}_t$ as (\ref{eq:smd-prediction-noupper-mu-gradient}).
       \State Compute predicted gradient \textcolor{black}{$\bar{\boldsymbol{g}}^{\boldsymbol{\mu}}_{t+1}$} for $\boldsymbol{\mu}$ as (\ref{eq:smd-prediction-noupper-mu-gradient-predicted}).
       \State Compute learning rate for $\boldsymbol{\mu}-$side:
       \begin{equation}
           \eta_t^{\mu} = \frac{\sqrt{2}}{2} \cdot \frac{\sqrt{\ln(N)}}{\sqrt{\sum_{i=1}^t \|\tilde{\boldsymbol{g}}^{\boldsymbol{\mu}}_i - \bar{\boldsymbol{g}}^{\boldsymbol{\mu}}_i\|_{\infty}^2}}.
           \label{eq:smd-prediction-noupper-mu-learning-rate}
       \end{equation}
       \State Update $\boldsymbol{\mu}_{t+1}$: $\forall \ell \in [N]$, $\mu_{t+1,\ell} =$
       \begin{equation}
           \frac{\mu_{t,\ell} \exp(-\eta_t^{\mu}(\tilde{g}_{t,\ell}^{\mu} - \bar{g}_{t,\ell}^{\mu}+\bar{g}_{t+1,\ell}^{\mu}))}{\sum_{\ell' = 1}^{N} \mu_{t,\ell'} \exp(-\eta_t^{\mu}(\tilde{g}_{t,\ell'}^{\mu} - \bar{g}_{t,\ell'}^{\mu}+\bar{g}_{t+1,\ell'}^{\mu}))}.
           \label{eq:smd-prediction-noupper-mu-update}
       \end{equation}
       \EndFor
       \State Compute $\bar{\boldsymbol{v}} = \frac{1}{T} \sum_{t \in [T]} \boldsymbol{v}_t$, $\bar{\boldsymbol{\mu}} = \frac{1}{T} \sum_{t \in [T]} \boldsymbol{\mu}_t$.
       \State Compute $\bar{\pi}$ such that
       \begin{equation*}
           \bar{\pi}_{(i, a)} = \frac{\bar{\mu}_{i, a}}{\sum_{a'\in {\cal A}_i} \bar{\mu}_{i, a'}}.
       \end{equation*}
       \State Output $\bar{\pi}$.
	\end{algorithmic}
	\label{alg:minimax-mirror-noupper}
\end{algorithm}

\subsection{Gradient Estimators}

\label{sec:alg-noupper-gradient-estimators}

The minimax optimization procedure requires knowledge of the gradients with respect to $\boldsymbol{v}$ and negative gradients with respect to $\boldsymbol{\mu}$ of $f(\cdot,\cdot)$. Specifically, for each pair of $(\boldsymbol{v},\boldsymbol{\mu})$, the following gradients are needed: For $\boldsymbol{v}-$side,
\begin{equation*}
    \boldsymbol{g}^{\boldsymbol{v}}(\boldsymbol{v},\boldsymbol{\mu}) = \nabla_{\boldsymbol{v}} f(\boldsymbol{v},\boldsymbol{\mu})=(1-\gamma) \boldsymbol{q} + \boldsymbol{\mu}^{\top} (\gamma \boldsymbol{\text{P}} - \hat{\boldsymbol{\text{I}}}),
\end{equation*}
and for $\boldsymbol{\mu}-$side,
\begin{equation*}
    \boldsymbol{g}^{\boldsymbol{\mu}}(\boldsymbol{v},\boldsymbol{\mu}) = - \nabla_{\boldsymbol{\mu}} f(\boldsymbol{v},\boldsymbol{\mu})= (\hat{\boldsymbol{\text{I}}} - \gamma \boldsymbol{\text{P}}) \boldsymbol{v} - \boldsymbol{\text{r}}.
\end{equation*}
However, these gradients cannot be directly obtained because $\boldsymbol{\text{P}}$ is unknown. Instead, we construct several estimators as proxies for the gradients in the algorithm.

\subsubsection{Stochastic Estimators}

We construct \textit{stochastic} estimators $\tilde{\boldsymbol{g}}^{\boldsymbol{v}}_t, \tilde{\boldsymbol{g}}^{\boldsymbol{\mu}}_t$ for gradients for $\boldsymbol{v}$, $\boldsymbol{\mu}$ through sampling. For $\boldsymbol{v}$-side, we construct the following stochastic gradient:
\begin{equation}
    \begin{aligned}
    & \text{Sample $(i,a) \sim \mu_{t,(i,a)}$, $j \sim p(j|i,a)$, $i'  \sim q_i$.} \\
    & \text{Compute} \\
    & \tilde{\boldsymbol{g}}^{\boldsymbol{v}}_t = \tilde{\boldsymbol{g}}^{\boldsymbol{v}}(\boldsymbol{v}_t,\boldsymbol{\mu}_t) = (1 - \gamma) \boldsymbol{e}_{i'} + \gamma \boldsymbol{e}_{j} - \boldsymbol{e}_i.
    \end{aligned}
    \label{eq:smd-prediction-noupper-v-gradient}
\end{equation}
This stochastic estimator is unbiased according to the following lemma.

\begin{lemma}{(Lemma 3 in \cite{jin2020efficiently})}
    The stochastic gradient $\tilde{\boldsymbol{g}}_t^{\boldsymbol{v}}$ for $\boldsymbol{v}$ satisfies $\mathbb{E}[\tilde{\boldsymbol{g}}_t^{\boldsymbol{v}}] = (1-\gamma) \boldsymbol{q} + \boldsymbol{\mu}_t^{\top} (\gamma \boldsymbol{\text{P}} - \hat{\boldsymbol{\text{I}}}) = \nabla_{\boldsymbol{v}} f(\boldsymbol{v}_t,\boldsymbol{\mu}_t)$. 
    \label{lem:property-gradient-v-noupper-unbiase}
\end{lemma}

For $\boldsymbol{\mu}$-side, we construct the following stochastic gradient:
\begin{equation}
    \begin{aligned}
        & \text{Sample $(i,a)$ from $ \mathcal{N}$ uniformly, $j \sim p(j|i,a)$.}\\
        & \text{Collect $z = (i,a,j) \in \mathcal{Z}$.} \\
        & \text{Compute $\tilde{\boldsymbol{g}}^{\boldsymbol{\mu}}_t = \tilde{\boldsymbol{g}}^{\boldsymbol{\mu}}(\boldsymbol{v}_t,\boldsymbol{\mu}_t) =$} \\
        & \frac{1}{t} \sum_{z = (i,j,a) \in \mathcal{Z}} N ( v_{t,i} - \gamma v_{t,j} - \text{r}_{i,a}) \boldsymbol{e}_{i,a}.
    \end{aligned}
    \label{eq:smd-prediction-noupper-mu-gradient}
\end{equation}

This stochastic estimator is unbiased as well. More importantly, its variance decreases over time by reusing past samples and averaging, ensuring greater stability and better control of the error bound in solving the minimax program.

\begin{lemma}
    The stochastic gradient $\tilde{\boldsymbol{g}}_t^{\boldsymbol{\mu}}$ for $\boldsymbol{v}$ satisfies $\mathbb{E}[\tilde{\boldsymbol{g}}_t^{\boldsymbol{\mu}}] = (\hat{\boldsymbol{\text{I}}} - \gamma \boldsymbol{\text{P}}) \boldsymbol{v}_t - \boldsymbol{\text{r}} = - \nabla_{\boldsymbol{\mu}} f(\boldsymbol{v}_t,\boldsymbol{\mu}_t) $ and $\mathbb{E}[\|\tilde{\boldsymbol{g}}_t^{\boldsymbol{\mu}}+\nabla_{\boldsymbol{\mu}} f(\boldsymbol{v}_t,\boldsymbol{\mu}_t) \|_{\infty}^2] \le 9N^2(1-\gamma)^{-2}/t$.
    \label{lem:property-gradient-mu-noupper-unbiase}
\end{lemma}

Lemma \ref{lem:property-gradient-v-noupper-unbiase}, \ref{lem:property-gradient-mu-noupper-unbiase} are proved in Appendix \ref{sec:app-pf-gradient-vmu-noupper}. The proof for Lemma \ref{lem:property-gradient-mu-noupper-unbiase} is inspired by Lemma 4 in \cite{jin2020efficiently}.


\subsubsection{Predicted Estimators}

Motivated by the idea of optimistic mirror descent in optimization and online learning \cite{rakhlin2013online,joulani2017modular,orabona2019modern}, we construct the following \textit{predicted} gradient as a proxy of the next-step gradient to facilitate updating $\boldsymbol{\mu}_t$ in (\ref{eq:smd-prediction-noupper-mu-update}). The predicted gradient is defined as follows
\textcolor{black}{
\begin{equation}
    \bar{\boldsymbol{g}}_{t+1}^{\boldsymbol{\mu}} = (\hat{\boldsymbol{\text{I}}} - \gamma \hat{\boldsymbol{\text{P}}}) \boldsymbol{v}_{t+1} - \boldsymbol{\text{r}}.
   \label{eq:smd-prediction-noupper-mu-gradient-predicted}
\end{equation}
}
This predicted gradient provides additional information to the algorithm, and its usefulness depends on the accuracy of $\hat{\boldsymbol{\text{P}}}$. Ideally, the true next-step gradient is $\boldsymbol{g}^{\boldsymbol{\mu}}(\boldsymbol{v}_{t+1},\boldsymbol{\mu}_{t+1}) = (\hat{\boldsymbol{\text{I}}} - \gamma \boldsymbol{\text{P}}) \boldsymbol{v}_{t+1} - \boldsymbol{\text{r}}$. Thus, the performance of optimistic mirror descent improves significantly when $\hat{\boldsymbol{\text{P}}}$ is sufficiently close to $\boldsymbol{\text{P}}$.

\subsection{Algorithm Description}

\label{sec:alg-noupper-description}

Our algorithm is presented in Algorithm \ref{alg:minimax-mirror-noupper}. Algorithm \ref{alg:minimax-mirror-noupper} consists of two parts: Line 2 to Line 11 represents the procedure of solving minimax problem (\ref{eq:lp-minimax-bellman-dmdp}), and Line 12 corresponds to the process of translating a feasible and approximately solution of (\ref{eq:lp-minimax-bellman-dmdp}) to a policy.

In order to leverage the potential benefit from the prediction, we develop a novel algorithmic framework to solve the minimax problem (\ref{eq:lp-minimax-bellman-dmdp}) iteratively. 
\textcolor{black}{The optimization procedure consists of $T$ steps.} During the optimization procedure, at step $t$, we compute the stochastic gradient estimators $\tilde{\boldsymbol{g}}^{\boldsymbol{v}}_t$ in Line 3 for $\boldsymbol{v}-$side and $\tilde{\boldsymbol{g}}^{\boldsymbol{\mu}}_t$ in Line 6 for $\boldsymbol{\mu}-$side through sampling. 
For the $\boldsymbol{v}$ side, we update $\boldsymbol{v}_t$ via mirror descent in (\ref{eq:smd-prediction-noupper-v-update}). The learning rate $\eta_t^{v}$ defined in (\ref{eq:smd-prediction-noupper-v-learning-rate}) is carefully designed such that \textcolor{black}{it is agnostic to the length of learning horizon $T$ and desired accuracy level $\epsilon$.} 
For the $\boldsymbol{\mu}$ side, additionally we construct a \textit{predicted} gradient in (\ref{eq:smd-prediction-noupper-mu-gradient-predicted}) using prediction transition matrix $\hat{\boldsymbol{\text{P}}}$. Then we update $\boldsymbol{\mu}_t$ via optimistic mirror descent in (\ref{eq:smd-prediction-noupper-mu-update}).
Similarly, the learning rate $\eta_t^{\mu}$ defined in (\ref{eq:smd-prediction-noupper-v-learning-rate}) is designed to \textcolor{black}{be agnostic not only to $T$ and $\epsilon$}, but also to the prediction error $\text{Dist}$.
\textcolor{black}{Our approach novelly integrates two gradient sources for updating $\boldsymbol{\mu}$: one from sampling and the other derived from predictions. Prediction-based gradients act as optimistic estimators, guiding the algorithm in regions where the predictions are accurate, thereby reducing exploration, while sample-based gradients provide unbiased updates, ensuring robustness.}

\textcolor{black}{Notably, we remark that our algorithm is parameter-free, requiring no prior knowledge of the desired accuracy level $\epsilon$ and prediction error $\text{Dist}$.} This yields greater practicability in implementation and applications.

Finally, in Line 12, we adapt the approach in \cite{jin2020efficiently} to convert an approximately optimal minimax solution into an approximately optimal policy.

\subsection{Analysis for Sample Complexity}

\label{sec:alg-noupper-analysis-sample-complexity}

Before proceeding, we introduce a performance metric to evaluate the quality of any feasible solution for minimax problem (\ref{eq:lp-minimax-bellman-dmdp}):
\begin{definition}
    For minimax problem (\ref{eq:lp-minimax-bellman-dmdp}), we define its \textit{duality gap} at a given pair of feasible solution $(\boldsymbol{v},\boldsymbol{\mu})$ as
    \begin{equation*}
        \text{GAP}(\boldsymbol{v},\boldsymbol{\mu}) = \max_{\boldsymbol{\mu}' \in \mathcal{U}} f(\boldsymbol{v},\boldsymbol{\mu}') - \min_{\boldsymbol{v}' \in \mathcal{V}} f(\boldsymbol{v}',\boldsymbol{\mu}).
    \end{equation*}
    Moreover, an $\epsilon$-optimal solution of the minimax problem (\ref{eq:lp-minimax-bellman-dmdp}) is a pair of feasible solution $(\boldsymbol{v}^{(\epsilon)},\boldsymbol{\mu}^{(\epsilon)})$ such that
\begin{equation*}
    \text{GAP}(\boldsymbol{v}^{(\epsilon)},\boldsymbol{\mu}^{(\epsilon)}) \le \epsilon.
\end{equation*}
\end{definition}

We are now ready to present the accuracy of the output $(\bar{\boldsymbol{v}},\bar{\boldsymbol{\mu}})$ in Line 11, in terms of the duality gap.
\begin{lemma}
    Given minimax problem (\ref{eq:lp-minimax-bellman-dmdp}), the solution $(\bar{\boldsymbol{v}},\bar{\boldsymbol{\mu}})$ \textcolor{black}{returned by OpPMD} satisfies
    \begin{equation*}
        \mathbb{E}[\text{GAP}(\bar{\boldsymbol{v}},\bar{\boldsymbol{\mu}})] \le \text{Err}_{v} + \text{Err}_{\mu,1} + \text{Err}_{\mu,2},
    \end{equation*}
    where
    \begin{equation*}
        \begin{aligned}
            & \text{Err}_{v} = 3 (1-\gamma)^{-1}\sqrt{|\mathcal{S}|} \cdot \sqrt{\frac{1}{T}} ,\\
            & \text{Err}_{\mu,1} = 3 \gamma (1-\gamma)^{-1}\sqrt{N} \min \left\{1,\text{Dist} \right\} \cdot \sqrt{\frac{1}{T}},\\
            & \text{Err}_{\mu,2} = 9 \sqrt{2} \frac{(1-\gamma)^{-1}N \ln(T)}{T}.
        \end{aligned}
    \end{equation*}
    The expectation is taken over the randomness of sampling procedure \textcolor{black}{in OpPMD.}
    \label{lem:minimax-error-noupper}
\end{lemma}

Lemma \ref{lem:minimax-error-noupper} is proved in Appendix \ref{sec:app-pf-minimax-error-noupper}. Particularly, $\text{Err}_{v}$ arises from the error in updating $\boldsymbol{v}$. $\text{Err}_{\mu,1}$ comes from the error of updating $\boldsymbol{\mu}$. $\text{Err}_{\mu,1}$ can be potentially improved whenever the prediction error $\text{Dist}$ is small. In other words, a possibly accurate prediction $\hat{\boldsymbol{\text{P}}}$ can accelerate convergence through the concept of optimistic mirror descent.
$\text{Err}_{\mu,2}$ is due to the variance of the stochastic gradient of $\boldsymbol{\mu}$.

\textcolor{black}{We define a pair of feasible solution $(\boldsymbol{v}^{(\epsilon)},\boldsymbol{\mu}^{(\epsilon)})$ is an \textit{expected} $\epsilon$-optimal solution of (\ref{eq:lp-minimax-bellman-dmdp}) if $\mathbb{E}[\text{GAP}(\boldsymbol{v}^{(\epsilon)},\boldsymbol{\mu}^{(\epsilon)})] \le \epsilon$. A policy $\pi_i^{(\epsilon)}$ is an \textit{expected} $\epsilon$-optimal policy with respect to $\boldsymbol{q}$ if $ v^*(\boldsymbol{q}) \le \mathbb{E}[v^{\boldsymbol{\pi}^{(\epsilon)}}(\boldsymbol{q})]  + \epsilon$. The expectations are taken over the randomness in the generation process of $(\boldsymbol{v}^{(\epsilon)},\boldsymbol{\mu}^{(\epsilon)})$ and $\pi_i^{(\epsilon)}$. By translating an expected approximately optimal solution into an expected approximately optimal policy, we derive the following sample complexity bound, which is proved in \ref{sec:app-pf-sample-complexity}.}




\begin{theorem}
    Given an DMDP model $\mathcal{M}$, let $\epsilon \in (0,1)$, OpPMD in Algorithm \ref{alg:minimax-mirror-noupper} can construct an expected $\epsilon$-optimal policy $\pi^{(\epsilon)}$ with prediction transition matrix $\hat{\boldsymbol{\text{P}}}$ with sample complexity 
    \begin{equation*}
        \tilde{O} \left (\max\left \{T_{v},T_{\mu,1},T_{\mu,2} \right \} \right),
    \end{equation*}
    where
    \begin{equation*}
        \begin{aligned}
            &T_{v} = (1-\gamma)^{-4} |\mathcal{S}| \epsilon^{-2}, \\
            & T_{\mu,1} = (1-\gamma)^{-4} N \min \left\{1,\text{Dist}^2 \right\} \epsilon^{-2}, \\
            & T_{\mu,2} = (1-\gamma)^{-2} N \epsilon^{-1}.
        \end{aligned}
    \end{equation*}
    \label{thm:sample-complexity-noupper}
\end{theorem}


We make several observations regarding the sample complexity in Theorem \ref{thm:sample-complexity-noupper}. First, this bound is uniformly better than $\tilde{O}((1-\gamma)^{-4}N\epsilon^{-2})$, the state-of-the-art result achieved by primal-dual algorithms \cite{jin2020efficiently}. Specifically, $T_{v}$, $T_{\mu,1}$, $T_{\mu,2}$ are all \textcolor{black}{less} than or equal to $\tilde{O}((1-\gamma)^{-4}N\epsilon^{-2})$. Since our approach is also a primal-dual method, this improvement highlights the value of leveraging predictions of the latent transition matrix, even when the prediction error is unknown. Second, when the prediction error is large (e.g. $\text{Dist} > 1$), 
$T_{\mu,1} = (1-\gamma)^{-4}N\epsilon^{-2}$, indicating that the bound degenerates to the best achievable result without using predictions. This demonstrates the robustness of our algorithm, as it can adaptively handle cases of poor predictions without negatively impacting overall performance. Intuitively, when the prediction is misleading with a large error, we should ignore it and learn from scratch. Finally, when $\text{Dist}^2 \le O((1-\gamma)^2\epsilon)$, our bound achieves its potentially best value $\tilde{O}(\max\{(1-\gamma)^{-4} |\mathcal{S}| \epsilon^{-2},(1-\gamma)^{-2} N \epsilon^{-1}\})$. This shows the potential for improvement brought by our approach when the prediction is sufficiently accurate.


\subsection{Sketch Proof for Lemma \ref{lem:minimax-error-noupper}}

\label{sec:alg-noupper-sketch-proof-minimax}

In this subsection, we provide a sketch proof for Lemma \ref{lem:minimax-error-noupper}. It is equivalent to prove a bound for 
$\mathbb{E}[f(\bar{\boldsymbol{v}},\boldsymbol{\mu}) - f(\boldsymbol{v},\bar{\boldsymbol{\mu}})]$ for any $\boldsymbol{v} \in \mathcal{V}$, $\boldsymbol{\mu} \in \mathcal{U}$, since 
\begin{equation*}
\begin{aligned}
\mathbb{E}[\text{GAP}(\bar{\boldsymbol{v}},\bar{\boldsymbol{\mu}})] &= \mathbb{E}[\max_{\boldsymbol{\mu}'} f(\bar{\boldsymbol{v}},\boldsymbol{\mu}') - \min_{\boldsymbol{v}'} f(\boldsymbol{v}',\bar{\boldsymbol{\mu}})] \\
&= \mathbb{E}[f(\bar{\boldsymbol{v}},\bar{\boldsymbol{\mu}}') - f(\bar{\boldsymbol{v}}',\bar{\boldsymbol{\mu}})],
\end{aligned}
\end{equation*}
$\bar{\boldsymbol{\mu}}' \in \mathop{\arg\max}_{\boldsymbol{\mu} \in \mathcal{U}} f(\bar{\boldsymbol{v}},\boldsymbol{\mu})$,  $\bar{\boldsymbol{v}}' \in \mathop{\arg\max}_{\boldsymbol{v} \in \mathcal{V}} f(\boldsymbol{v},\bar{\boldsymbol{\mu}})$. We can decompose it as follows:
\begin{subequations}
    \begin{align}
        & \mathbb{E} \left[ f(\bar{\boldsymbol{v}},\boldsymbol{\mu}) - f(\boldsymbol{v},\bar{\boldsymbol{\mu}})\right] \nonumber \\
        \le & \frac{1}{T} \mathbb{E}\left [ \sum_{t=1}^T \tilde{\boldsymbol{g}}^{\boldsymbol{\mu}}(\boldsymbol{v}_t,\boldsymbol{\mu}_t)^{\top} \boldsymbol{\mu}_t - \sum_{t=1}^T \tilde{\boldsymbol{g}}^{\boldsymbol{\mu}}(\boldsymbol{v}_t,\boldsymbol{\mu}_t)^{\top} \boldsymbol{\mu} \right] \label{eq:skproof-minimax-error-decompose-mu} \\
        + & \frac{1}{T}\mathbb{E}\left [ \sum_{t=1}^T \tilde{\boldsymbol{g}}^{\boldsymbol{v}}(\boldsymbol{v}_t,\boldsymbol{\mu}_t)^{\top} \boldsymbol{v}_t - \sum_{t=1}^T \tilde{\boldsymbol{g}}^{\boldsymbol{v}}(\boldsymbol{v}_t,\boldsymbol{\mu}_t)^{\top} \boldsymbol{v}\right]. \label{eq:skproof-minimax-error-decompose-v}
        \end{align}
    \label{eq:skproof-minimax-error-decompose}
\end{subequations}
(\ref{eq:skproof-minimax-error-decompose-mu}) reflects the error from updating $\boldsymbol{\mu}$ and (\ref{eq:skproof-minimax-error-decompose-v}) reflects the error from updating $\boldsymbol{v}$. We analyze these two terms respectively. For (\ref{eq:skproof-minimax-error-decompose-mu}), $T \cdot (\ref{eq:skproof-minimax-error-decompose-mu})$
\begin{subequations}
    \begin{align}
        &= \mathbb{E} \left[ \frac{\ln(N)}{\eta_T^{\mu}} + \frac{1}{2} \sum_{t=1}^T \eta_t^{\mu} \|\tilde{\boldsymbol{g}}_t^{\boldsymbol{\mu}}  - \bar{\boldsymbol{g}}_t^{\boldsymbol{\mu}}\|_{\infty}^2 \right] \label{eq:skproof-minimax-error-mu-side-a} \\
        & = O \left (\sqrt{\ln(N)} \right ) \cdot \left( \mathbb{E}\left[ \sqrt{\sum_{t=1}^T \|\tilde{\boldsymbol{g}}_t^{\boldsymbol{\mu}}  - \bar{\boldsymbol{g}}_t^{\boldsymbol{\mu}}\|_{\infty}^2} \right] \right.  \nonumber \\
    & +  \left.   \mathbb{E} \left[ \sum_{t=1}^T \frac{\|\tilde{\boldsymbol{g}}_t^{\boldsymbol{\mu}}  - \bar{\boldsymbol{g}}_t^{\boldsymbol{\mu}}\|_{\infty}^2}{\sqrt{\sum_{i=1}^t \|\tilde{\boldsymbol{g}}^{\boldsymbol{\mu}}_i - \bar{\boldsymbol{g}}^{\boldsymbol{\mu}}_i\|_{\infty}^2}} \right ] \right) \label{eq:skproof-minimax-error-mu-side-b} \\
    & \le  O\left(\sqrt{\ln(N)} \right) \cdot \mathbb{E} \left[\sqrt{\sum_{t=1}^T \|\tilde{\boldsymbol{g}}_t^{\boldsymbol{\mu}}  - \bar{\boldsymbol{g}}_t^{\boldsymbol{\mu}}\|_{\infty}^2} \right] \label{eq:skproof-minimax-error-mu-side-c}\\
    & \le  O \left( \sqrt{\ln(N)}  \right) \cdot \left ( \sqrt{\mathbb{E} \left[\sum_{t=1}^T  \|\tilde{\boldsymbol{g}}_t^{\boldsymbol{\mu}}  - \boldsymbol{g}^{\boldsymbol{\mu}}(\boldsymbol{v}_t,\boldsymbol{\mu}_t) \|_{\infty}^2\right]} \right. \nonumber \\ 
    & + \left. \sqrt{\mathbb{E} \left[\sum_{t=1}^T  \| \boldsymbol{g}^{\boldsymbol{\mu}}(\boldsymbol{v}_t,\boldsymbol{\mu}_t) -\bar{\boldsymbol{g}}_t^{\boldsymbol{\mu}}\|_{\infty}^2\right]}  \right) \label{eq:skproof-minimax-error-mu-side-d} \\
    & \le T \cdot \tilde{O} \left( \text{Err}_{\mu,2} +  \text{Err}_{\mu,1}\right). \nonumber
    \end{align}
    \label{eq:skproof-minimax-error-mu-side}
\end{subequations}
(\ref{eq:skproof-minimax-error-mu-side-a}) comes from the performance guarantee for optimistic (online) mirror descent. (\ref{eq:skproof-minimax-error-mu-side-b}) comes from the definition of learning rate $\eta_t^{\mu}$, (\ref{eq:skproof-minimax-error-mu-side-c}) comes from Lemma 4.13 in \cite{orabona2019modern}. (\ref{eq:skproof-minimax-error-mu-side-d}) comes from Jensen inequality and triangle inequality. We remark that (\ref{eq:skproof-minimax-error-mu-side-b}) and (\ref{eq:skproof-minimax-error-mu-side-c}) demonstrates how the adaptive learning rates work. Similarly, for (\ref{eq:skproof-minimax-error-decompose-v}), $T \cdot (\ref{eq:skproof-minimax-error-decompose-v})$
\begin{subequations}
    \begin{align}
        & = \mathbb{E} \left [ \frac{|\mathcal{S}|(1-\gamma)^{-2}}{\eta_T^v} + \frac{1}{2} \sum_{t=1}^T \eta_t^v \|\tilde{\boldsymbol{g}}^{\boldsymbol{v}}(\boldsymbol{v}_t,\boldsymbol{\mu}_t)\|_2^2\right] \label{eq:skproof-minimax-error-v-side-a}\\
        & = O \left(\sqrt{|\mathcal{S}|} (1-\gamma)^{-1} \right) \cdot \left( \mathbb{E} \left [\sqrt{\sum_{t=1}^T\|\tilde{\boldsymbol{g}}^{\boldsymbol{v}}(\boldsymbol{v}_t,\boldsymbol{\mu}_t)\|_2^2} \right]\right. \nonumber \\
        & + \left. \mathbb{E} \left[\sum_{t=1}^T \frac{\|\tilde{\boldsymbol{g}}^{\boldsymbol{v}}(\boldsymbol{v}_t,\boldsymbol{\mu}_t)\|_2^2}{\sqrt{\sum_{i=1}^t \|\tilde{\boldsymbol{g}}^{\boldsymbol{v}}(\boldsymbol{v}_i,\boldsymbol{\mu}_i)\|_2^2}}  \right]\right) \label{eq:skproof-minimax-error-v-side-b} \\
        & \le O \left( \sqrt{|\mathcal{S}|} (1-\gamma)^{-1} \right) \cdot \mathbb{E} \left[\sqrt{\sum_{t=1}^T\|\tilde{\boldsymbol{g}}^{\boldsymbol{v}}(\boldsymbol{v}_t,\boldsymbol{\mu}_t)\|_2^2} \right ] \label{eq:skproof-minimax-error-v-side-c} \\
        & \le  O \left( \sqrt{|\mathcal{S}|} (1-\gamma)^{-1} \right) \cdot  \sqrt{ \mathbb{E} \left[\sum_{t=1}^T\|\tilde{\boldsymbol{g}}^{\boldsymbol{v}}(\boldsymbol{v}_t,\boldsymbol{\mu}_t)\|_2^2 \right]} \nonumber \\
        & \le T \cdot O \left( \text{Err}_{v} \right). \nonumber
    \end{align}
    \label{eq:skproof-minimax-error-v-side}
\end{subequations}
(\ref{eq:skproof-minimax-error-v-side-a}) comes from the performance guarantee for mirror descent, (\ref{eq:skproof-minimax-error-v-side-b}) comes from the definition of learning rate $\eta_t^v$, (\ref{eq:skproof-minimax-error-v-side-c}) comes from Lemma 4.13 in \cite{orabona2019modern}. Altogether, the Theorem is proved.





\section{Numerical Experiments}

\label{sec:numerical}

In this section, we provide numerical experiments to validate our approach. We consider a simple MDP instance. The details of the instance and experimental settings (e.g., $\gamma$, $\epsilon$, $\boldsymbol{q}$) are provided in Appendix \ref{sec:app-details-numerical}. We evaluate and compare three algorithms in this environment: (1) OpPMD-AC: Algorithm \ref{alg:minimax-mirror-noupper} with accurate prediction input, i.e. $\hat{\boldsymbol{\text{P}}} = \boldsymbol{\text{P}}$. (2) OpPMD-NAC: Algorithm \ref{alg:minimax-mirror-noupper} with inaccurate prediction, i.e. $\hat{\boldsymbol{\text{P}}} \ne \boldsymbol{\text{P}}$ and $\text{Dist}(\hat{\boldsymbol{\text{P}}} , \boldsymbol{\text{P}}) > 1$. The definition of $\hat{\boldsymbol{\text{P}}}$ for OpPMD-NAC is also provided in Appendix \ref{sec:app-details-numerical}. (3) SMD-DMDP-JINSID: Algorithm 1 in \cite{jin2020efficiently}, designed to solve DMDPs without any predictions.

\begin{figure}[ht]
\vskip -0.1in
\begin{center}
\centerline{\includegraphics[width=\columnwidth]{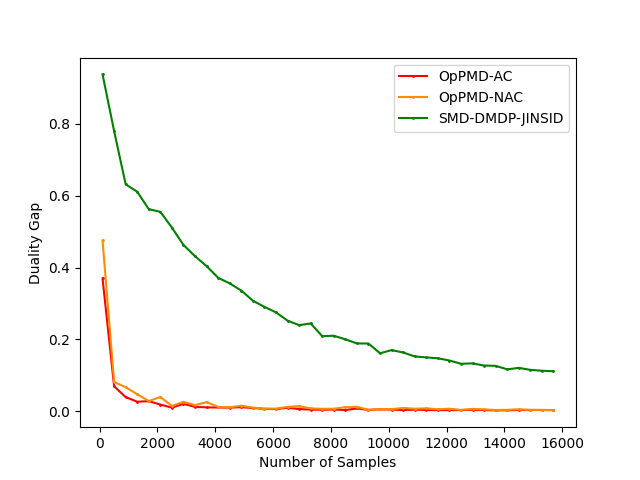}}
\vskip -0.15in
\caption{Duality Gap}
\label{fig:mdp-duality-gap}
\end{center}
\vskip -0.3in
\end{figure}

\begin{figure}[ht]
\vskip -0.1in
\begin{center}
\centerline{\includegraphics[width=\columnwidth]{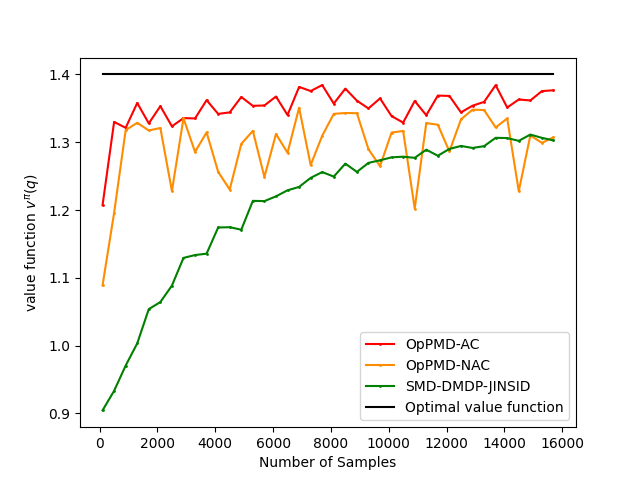}}
\vskip -0.15in
\caption{Value Function}
\label{fig:mdp-value-function}
\end{center}
\vskip -0.3in
\end{figure}

Figure \ref{fig:mdp-duality-gap} plots the Duality Gap ($\text{GAP}(\bar{\mu}, \bar{v})$), and Figure \ref{fig:mdp-value-function} presents the value function. It is important to note that the superiority in numerical performance of OpPMD-AC does not imply that our algorithm is uniformly superior to other benchmarks, like \cite{jin2020efficiently}. This performance advantage arises because OpPMD-AC utilizes accurate predictions of the transition matrix, whereas the other algorithms do not. The results highlight the potential benefits of having sufficiently accurate predictions of the transition matrix. Furthermore, they emphasize the importance of a well-designed algorithm, such as our OpPMD, in effectively leveraging these predictions to improve performance.


\bibliography{mdp}

\newpage
\appendix
\onecolumn

\section{More Detailed Discussions on Existing Works}

\subsection{More Discussions on How to Construct Prediction Transition Matrix $\hat{\boldsymbol{\text{P}}}$}

\label{sec:app-discuss-prediction-P}

In many applications of MDPs and Reinforcement Learning, the prediction transition matrix $\hat{\boldsymbol{\text{P}}}$ can be readily available or can be directly constructed. The applications has been previously explored by \cite{golowich2022can}, which studies online tabular finite-horizon episodic MDPs with predictions in Q-value. Our framework expands the research landscape by (1) incorporating predictions to solve infinitie-horizon MDPs and improve sample complexity bounds, whereas prior work has primarily focused on regret, and (2) exploring predictions on the transition matrix rather than on the Q-value, which is often easier to obtain in practice. In the following, we explain the rationale behind our model.
Interesting readers can consult \cite{golowich2022can} for more discussions.


\textbf{Physics Simulation} There are several powerful physics simulation engines in robotics, such as MuJoCo \cite{todorov2012mujoco,lazaridis2020deep,tessler2019action,weng2022tianshou}, which allow us to collect huge amount of data from simulations. While such data may not be directly applicable to the decision-making process due to the dynamic changing environment, it can be used to construct various predictions, such as $\hat{\boldsymbol{\text{P}}}$ in our paper or Q-value in \cite{golowich2022can} to enhance learning algorithms. Several related works include \cite{zhao2020sim,peng2018sim,collins2021review}.

\textbf{Healthcare} In healthcare applications \cite{yu2021reinforcement}, extensive records from previous clinical trials capture treatments and outcomes from numerous patients. Although individuals differ, it is reasonable to assume that learning tasks exhibit similarities across patients. This allows for leveraging these records to construct predictions, such as transition matrices. There are some related work, such as \cite{chen2022data,mate2022field,liao2020personalized}. Additionally, various widely used health apps, such as Apple Health, MyFitnessPal, Sleep Cycle, collect vast amounts of user data that can contribute to predictive modeling and personalized insights.

\textbf{Transfer Learning and Multi-task Learning} Our framework for leveraging predictions to enhance the solving of DMDPs is related to transfer learning or multi-task learning. In the context of reinforcement learning (RL), these approaches aim to use knowledge from one RL task to improve learning efficiency and performance across several different but related RL tasks. This can significantly reduce the time, data or computational resources required to learn new tasks. The prediction matrix $\hat{\boldsymbol{\text{P}}}$ in our framework can be viewed as an estimation of the transition matrix from an earlier task, which can be utilized to accelerate learning future related tasks. Interesting readers can refer to \cite{taylor2009transfer,zhu2023transfer,hua2021learning,gamrian2019transfer} for further details.

\subsection{Comparing with \cite{jin2020efficiently}}

\label{sec:app-com-sidford-efficient}

\cite{jin2020efficiently} provides a unified framework for finding an expected $\epsilon$-optimal policy for both infinite DMDPs and AMDPs. Similar to our work, their approach is based on solving minimax problem (\ref{eq:lp-minimax-bellman-dmdp}) iteratively using stochastic mirror descent. Specifically, they design stochastic unbiased estimators for $\boldsymbol{v}$-, $\boldsymbol{\mu}$-sides, respectively, and iteratively update $\boldsymbol{v}$ and $\boldsymbol{\mu}$. Finally, for DMDPs they transform a $(1-\gamma) \epsilon$-optimal solution for (\ref{eq:lp-minimax-bellman-dmdp}) $(\boldsymbol{v}^{(\epsilon)},\boldsymbol{\mu}^{(\epsilon}))$, obtained from the minimax optimization process, into an $\epsilon-$optimal policy $\pi^{(\epsilon)}$. For DMDPs, their approach achieves a sample complexity bound of $\tilde{O}((1-\gamma)^{-4}N\epsilon^{-2})$, matching the lower bound up to a $(1-\gamma)^{-1}$ factor. In comparison, our work studies a different problem that we go a further step to see whether predictions of the transition matrix $\hat{\boldsymbol{\text{P}}}$ can enhance the process of computing an (approximately) optimal policy and improve the sample complexity bound. We make several improvements on \cite{jin2020efficiently}'s approach to make it adaptive into our setting. 

First, we construct additional \textit{predicted} gradients for the $\boldsymbol{\mu}$ side for time $t+1$ in (\ref{eq:smd-prediction-noupper-mu-gradient-predicted}), facilitating the update of $\boldsymbol{\mu}_t$ at time $t$. This allows us to benefit from the power of optimistic mirror descent, enhancing the algorithm performance.

Second, we carefully design our learning rates to eliminate the dependence on the prediction error $\text{Dist}$ and the desired accuracy level $\epsilon$. In \cite{jin2020efficiently}, the learning rates are defined as
\begin{equation*}
    \eta_t^v = \frac{\epsilon}{8},\qquad \eta_t^\mu = \frac{\epsilon}{36 ((1-\gamma)^{-2}+1)N},
\end{equation*}
which rely on the value of $\epsilon$. If we were to construct our learning rates similarly, the learning rates should be 
\begin{equation*}
    \eta_t^v = O (\epsilon),\qquad \eta_t^\mu = O \left( \frac{\epsilon}{N \gamma^2(1-\gamma)^{-2}\min \{1,\text{Dist}^2\} }\right),
\end{equation*}
which depend on both $\epsilon$ and prediction error $\text{Dist}$. To address this, we adopt a parameter-free approach using adaptive learning rates, relaxing the need to know these two values.

Third, we provide a slightly different approach to construct the stochastic estimators for the gradients on the $\boldsymbol{\mu}$-side. Our method includes a simple variance reduction technique to control the variance of the stochastic estimators, shrinking them as the sample size increases. This ensures better control in error analysis.

Altogether, while our work is inspired by the minimax approach proposed by \cite{jin2020efficiently}, we introduce essential and novel improvements over \cite{jin2020efficiently} to incorporate the black-box predictions into the process of computing (approximately) optimal policy.

\subsection{Comparing with (Online) Algorithm with Advice (Prediction)}

\label{sec:app-com-alg-advice}

Regarding the prediction model, our work aligns with the research stream of (Online) Algorithm with Black-box Advice (Prediction) \cite{mitzenmacher2022algorithms,purohit2018improving}. 
This line of research focuses on improving algorithm performance in specific problem settings by incorporating black-box advice (predictions) before the decision-making process.
\cite{mitzenmacher2022algorithms,purohit2018improving} define the notions of robustness and consistency, where performance metrics are based on balancing these two aspects in terms of competitive ratio.
Various specific online problems have been explored, including caching \cite{lykouris2021competitive,rohatgi2020near}, online resource allocation \cite{jiang2020online, balseiro2022single,golrezaei2023online}, online matching \cite{jin2022online}, online secretary \cite{antoniadis2020secretary,dutting2021secretaries}, and convex optimization \cite{christianson2022chasing}. For more references, see \url{https://algorithms-with-predictions.github.io/}. In comparison, our work is the first to investigate leveraging predictions to enhance learning MDPs. Moreover, unlike these studies, our work focuses on sample complexity bound rather than competitive ratio-based metrics.

Another related line of research considers online optimization with predictions, aiming to decrease regret with methods like optimistic mirror descent, where the predictions are provided sequentially. Full feedback models are studied in \cite{rakhlin2013online,rakhlin2013optimization,jadbabaie2015online}, the multi-armed bandit is studied in \cite{wei2018more}, and the contextual bandit is studied in \cite{wei2020taking}.
We draw inspiration from algorithmic techniques from this line of research, such as optimistic mirror descent, to effectively incorporate black-box predictions into our minimax optimization framework. However, we emphasize that these methods cannot be directly applied to our approach. Unlike dynamic predictions, we rely on a fixed and single prediction $\hat{\boldsymbol{\text{P}}}$ provided in advance of the decision-making process. Furthermore, in our algorithm, the feedback at each time step is stochastic and unbiased full feedback, differing from deterministic full feedback \cite{rakhlin2013online,rakhlin2013optimization,jadbabaie2015online} or bandit feedback \cite{wei2018more,wei2020taking}, necessitating a distinct analytical framework.




\section{Proof for Section \ref{sec:impossibility-result}}

\subsection{Notations and Auxiliary Results}

\textbf{Notations} Consider a fixed algorithm ALG and an DMDP instance $\mathcal{M}$ with transition matrix $\boldsymbol{\text{P}}$. We denote $\rho_{\mathcal{M},\text{ALG}}$ be the joint probability distribution on $\{(i_t,a_t,j_t)\}_{t=1}^T$ (where $j_t \sim p(\cdot|i_t,a_t)$), the trajectory under algorithm ALG on instance $\mathcal{M}$ with $T$ samples. For a $\sigma(\{(i_t,a_t,j_t)\}_{t=1}^T)-$measurable event $A$, we denote $\Pr_{\mathcal{M},\text{ALG}}(A)$ be the probability of $A$ under $\rho_{\mathcal{M},\text{ALG}}$. For a $\sigma(\{(i_t,a_t,j_t)\}_{t=1}^T)-$measurable random variable $X$, we denote $\mathbb{E}_{\mathcal{M},\text{ALG}}[X]$ as the expectation of $X$ under $\rho_{\mathcal{M},\text{ALG}}$. We denote $N_{T}(i,a)$ be the number of samples on state-action pair $(i,a)$ during the whole sampling horizon, satisfying $\sum_{(i,a)\in\mathcal{N}}N_{T}(i,a) = T$. For a fixed policy ALG, for simplicity, we abbreviate $\Pr_{\mathcal{M},\text{ALG}}(\cdot)$, $\mathbb{E}_{\mathcal{M},\text{ALG}}[\cdot]$ as $\Pr_{\mathcal{M}}(\cdot)$, $\mathbb{E}_{\mathcal{M}}[\cdot]$, respectively.

\textbf{Auxiliary Results} Now we provide some auxiliary results from \cite{lattimore2020bandit}.

\begin{theorem}{(Bretagnolle–Huber inequality, Theorem 14.2 in \cite{lattimore2020bandit})}
    Let $\mathbb{P}$, $\mathbb{Q}$ be probability measures on $(\Omega,\mathcal{F})$, and let $A \in \mathcal{F}$ be an arbitrary event. Then
    \begin{equation*}
        \Pr_{\mathbb{P}}(A) + \Pr_{\mathbb{Q}}(A^c) \ge \frac{1}{2} \exp \left(- \text{KL}(\mathbb{P},\mathbb{Q}) \right).
    \end{equation*}
    \label{thm:BH-inequality}
\end{theorem}

\begin{theorem}{(Divergence decomposition, Lemma 15.1 in \cite{lattimore2020bandit})}
    Consider two instances $\mathcal{M}_1$, $\mathcal{M}_2$ that share the same state space, same action space for each state, same number of samples $T$. For any sampling algorithm ALG,
    \begin{equation*}
        \text{KL}(\Pr_{\mathcal{M}_1},\Pr_{\mathcal{M}_2}) = \sum_{(i,a) \in \mathcal{N}} \mathbb{E}_{\mathcal{M}_1}[N_T(i,a)] \cdot \text{KL}(\Pr_{\mathcal{M}_1,(i,a)},\Pr_{\mathcal{M}_2,(i,a)}).
    \end{equation*}
    \label{thm:divergence-decomposition}
\end{theorem}

\subsection{Proof for Theorem \ref{thm:impossibility-result}}

\label{sec:app-pf-thm-impossibility}

This proof is inspried by the lower bound example provided by \cite{gheshlaghi2013minimax}. Throughout the proof we assume $\gamma \ge \frac{1}{3}$.

We define model class $\mathcal{I}_{m,n}$, with $m,n \ge 1$ and $mn > 1$: For any DMDP model $\mathcal{M} \in \mathcal{I}_{m,n}$,
\begin{itemize}
    \item State $\mathcal{S}$: $\mathcal{S}$ includes $m$ ``start'' states $\{i^{(\text{st})}_k\}_{k \in [m]}$, $mn$ ``middle'' states $\{i^{(\text{mi})}_{k,\ell}\}_{k \in [m], \ell \in [n]}$, and $mn$ ``end'' states $\{i^{(\text{en})}_{k,\ell}\}_{k \in [m], \ell \in [n]}$.
    \item Action $\mathcal{A}$: 
    \begin{itemize}
        \item ``start'' states $\{i^{(\text{st})}_k\}_{k \in [m]}$: For each $k \in [m]$, the state $i^{(\text{st})}_k$ can take $n$ actions $\{a_{\ell}\}_{\ell \in [n]}$. For each state action pair $(i^{(\text{st})}_k,a_{\ell})$, it receives reward 0 and transits to state $i^{(\text{mi})}_{k,\ell}$ with probability 1.
        \item ``middle'' states $\{i^{(\text{mi})}_{k,\ell}\}_{k \in [m], \ell \in [n]}$: For each $(k,\ell)$, the state $i^{(\text{mi})}_{k,\ell}$ can only take 1 action $a^{(\text{mi})}$. After taking this action, it receives reward 1 with probability 1. It transits to itself with probability $p_{\mathcal{M}}(k,\ell) \in [0,1]$, and with probability $1 - p_{\mathcal{M}}(k,\ell)$ transits to $i^{(\text{en})}_{k,\ell}$.
        \item ``end'' states $\{i^{(\text{en})}_{k,\ell}\}_{k \in [m], \ell \in [n]}$: For each $(k,\ell)$, the state $i^{(\text{en})}_{k,\ell}$ can only take 1 action $a^{(\text{en})}$, receiving reward 0 and transiting to itself with probablity 1.
    \end{itemize}
\end{itemize}
Note that the total number of state-action pairs $N = 3mn$. We define instance $\mathcal{M}_0 \in \mathcal{I}_{m,n}$ with transition matrix $\boldsymbol{\text{P}}_0$ and prediction matrix $\hat{\boldsymbol{\text{P}}} = \boldsymbol{\text{P}}_0$, which is defined as follows:
\begin{equation*}
    p_{\mathcal{M}_0}(k,\ell) = \begin{cases}
        p_0 + \Delta & (k,\ell) = (1,1), \\
        p_0 & \text{otherwise.}
    \end{cases}
\end{equation*}
where $p_0 = \frac{4 \gamma - 1}{3 \gamma}$, and $\Delta$ is to be determined later. For this model, the optimal value vector is
\begin{equation*}
    v^{*,(\text{en})}_{\mathcal{M}_0,k,\ell} = 0,\qquad v^{*,(\text{mi})}_{\mathcal{M}_0,k,\ell} = \begin{cases}
        \frac{1}{1-\gamma(p_0+\Delta)} & (k,\ell) = (1,1),\\
        \frac{1}{1-\gamma p_0} & \text{otherwise,}   
    \end{cases} \qquad  v^{*,(\text{st})}_{\mathcal{M}_0,k} = \gamma \max_{\ell \in [n]} v^{*,(\text{mi})}_{\mathcal{M}_0,k,\ell} .
\end{equation*}

For simplicity, we further assume that for any algorithm ALG that access to the knowledge of the knowledge of transition for ``start'' and ``end'' states, and only does not know about the knowledge of transition probability for ``middle'' states $\{p_{\mathcal{M}_0}(k,\ell)\}_{k \in [m], \ell \in [n]}$.
Now, we show that for any ALG, if it can be $(\epsilon,\delta)$-\textit{smart} ($\epsilon \in (0, \frac{1}{40} (1-\gamma)^{-1})$, $\delta \in (0, 0.24)$ on $\mathcal{M}_0$ with the help of prediction $\hat{\boldsymbol{\text{P}}}$ with $T$ satisfying
\begin{equation*}
    T \le \frac{1}{300}(1-\gamma)^{-3} (mn-1) \epsilon^{-2} \ln\left(\frac{1}{4.1\delta}\right),
\end{equation*}
then there exists $\mathcal{M}' \in \mathcal{I}_{m,n}$ such that ALG cannot be $(\epsilon,\delta)$-\textit{smart} on $\mathcal{M}' $ with a same $T$ and same prediction $\hat{\boldsymbol{\text{P}}}$.

We start from the follow definition for $N_T(\cdot,\cdot)$:
\begin{equation*}
    \sum_{k \in [m],\ell \in[n]}\mathbb{E}_{\mathcal{M}_0} [N_T(i^{(\text{mi})}_{k,\ell},a^{\text{(mi)}})] = T.
\end{equation*}
Then there exist $(k',\ell') \ne (1,1)$ such that
\begin{equation*}
    \mathbb{E}_{\mathcal{M}_0} [N_T(i^{(\text{mi})}_{k',\ell'},a^{\text{(mi)}})] \le \frac{T}{mn-1}.
\end{equation*}
Now, we construct $\mathcal{M}' \in \mathcal{I}_{m,n}$ as follows:
\begin{equation*}
    p_{\mathcal{M}'}(k,\ell) = \begin{cases}
        p_0 + \Delta & (k,\ell) = (1,1), \\
        p_0 + 2 \Delta & (k,\ell) = (k',\ell'),\\
        p_0 & \text{otherwise.}
    \end{cases}
\end{equation*}

Let $\Delta = \frac{5}{3} \cdot \frac{(1-\gamma)^2}{\gamma} \epsilon$, then we have
\begin{subequations}
    \begin{align}
    & \Pr_{\mathcal{M}_0,T} \left( \left\| v^{\hat{\pi}_T} - v^*_{\mathcal{M}_0} \right \|_{\infty}  >\epsilon \right) + \Pr_{\mathcal{M}',T} \left( \left\| v^{\hat{\pi}_T} - v^*_{\mathcal{M}'} \right \|_{\infty}  >\epsilon \right) \nonumber \\
    \ge  & \Pr_{\mathcal{M}_0,T} \left( \left\| v^{\hat{\pi}_T} - v^*_{\mathcal{M}_0} \right \|_{\infty}  >\epsilon \right) + \Pr_{\mathcal{M}',T} \left( \left\| v^{\hat{\pi}_T} - v^*_{\mathcal{M}_0} \right \|_{\infty} \le \epsilon \right) \label{eq:lower-derive-a} \\
    \ge & \frac{1}{2} \exp \left[-\text{KL}\left(\Pr_{\mathcal{M}_0,T},\Pr_{\mathcal{M}',T} \right) \right] \label{eq:lower-derive-b}\\
    = & \frac{1}{2} \exp \left[ - \mathbb{E}_{\mathcal{M}_0} [N_T(i^{(\text{mi})}_{k',\ell'},a^{\text{(mi)}})]\cdot\text{KL}\left(\text{Bern}\left( p_0\right),\text{Bern}\left( p_0 + 2 \Delta\right)\right) \right] \label{eq:lower-derive-c}\\
    \ge & \frac{1}{2} \exp \left[-\frac{1}{300} \cdot \frac{(1-\gamma)^{-3} (mn-1) \epsilon^{-2}}{mn-1} \cdot \frac{4 \Delta^2}{p_0(1-p_0)} \cdot \ln\left(\frac{1}{4.1\delta} \right)\right] \label{eq:lower-derive-d} \\
    \ge & \frac{1}{2} \exp \left[-\frac{1}{300} \cdot \frac{(1-\gamma)^{-3} (mn-1) \epsilon^{-2}}{mn-1} \cdot \frac{12 \Delta^2}{1-p_0}\cdot \ln\left(\frac{1}{4.1\delta}\right)\right]  \label{eq:lower-derive-e} \\
    \ge & \frac{1}{2} \exp\left(- \ln\left(\frac{1}{4.1\delta}\right)\right) > 2 \delta. \label{eq:lower-derive-f}
    \end{align}
    \label{eq:lower-derive}
\end{subequations}
where (\ref{eq:lower-derive-a}) comes from the following: by $\epsilon < \frac{1}{40} (1-\gamma)^{-1}$, we have $2 \Delta \le \frac{1 - \gamma}{12 \gamma}$, then
\begin{equation*}
    p_0 + 2 \Delta \le  \frac{4 \gamma-1}{3 \gamma} + \frac{1 - \gamma}{12 \gamma} = \frac{5\gamma - 1}{4 \gamma}, \qquad \Rightarrow \qquad  \frac{1}{1 - \gamma(p_0+2\Delta)} \ge \frac{4}{5} \frac{1}{1-\gamma}.
\end{equation*}
We can also derive $\frac{1}{1-\gamma p_0} = \frac{3}{4} \frac{1}{1 - \gamma}$. Therefore,
\begin{equation*}
    \begin{aligned}
    \frac{1}{1 - \gamma(p_0 + 2 \Delta)} - \frac{1}{1 - \gamma p_0} = \frac{2\gamma \Delta}{(1 - \gamma(p_0 + 2 \Delta))(1 - \gamma p_0)} \ge 2 \gamma \cdot \frac{5}{3} \frac{(1-\gamma)^2 \epsilon}{\gamma} \cdot \frac{4}{5} \frac{1}{1 - \gamma} \cdot \frac{3}{4} \frac{1}{1 - \gamma} = 2 \epsilon.
    \end{aligned}
\end{equation*}
This implies that the subset of events that perform ``bad'' at $\mathcal{M}'$ ($\left\| v^{\hat{\pi}_T} - v^*_{\mathcal{M}_0} \right \|_{\infty} \le \epsilon$) is a subset of events that perform ``well'' at $\mathcal{M}_0$ ($\left\| v^{\hat{\pi}_T} - v^*_{\mathcal{M}'} \right \|_{\infty}  >\epsilon$). Or in other words, here the difference between $\mathcal{M}_0$ and $\mathcal{M}'$ is large enough that when it can perform well or either of the instances and can not on both. (\ref{eq:lower-derive-b}) comes from Bretagnolle–Huber inequality (Theorem \ref{thm:BH-inequality}). (\ref{eq:lower-derive-c}) comes from Divergence Decomposition (Theorem \ref{thm:divergence-decomposition}). (\ref{eq:lower-derive-d}) comes from following:
\begin{equation*}
    \begin{aligned}
    \text{KL}\left(\text{Bern}\left( p_0\right),\text{Bern}\left( p_0 + 2 \Delta\right)\right) & = p_0 \ln \left(\frac{p_0}{p_0 + 2 \Delta} \right) + (1 -p_0) \ln \left(\frac{1 - p_0}{1 - p_0 - 2 \Delta} \right) \\
    & = - p_0 \ln \left(\frac{p_0 + 2 \Delta}{p_0 } \right) - (1 -p_0) \ln \left(\frac{1 - p_0 - 2\Delta}{1 - p_0 } \right) \\
    & \le p_0 \left (\frac{4\Delta^2}{p_0} - \frac{4\Delta}{p_0}\right) + (1-p_0) \left(\frac{4\Delta^2}{(1-p_0)^2} + \frac{2 \Delta}{1 - p_0}\right) \\
    &  \le \frac{4\Delta^2}{p_0} + \frac{4 \Delta^2}{ 1- p_0} = \frac{4 \Delta^2}{p_0(1 - p_0)},
    \end{aligned}
\end{equation*}
where the first inequality comes from the following two facts:
\begin{equation*}
    \ln (1 + x) \ge x - \frac{x^2}{2}, \qquad \forall x \in [0,1], \qquad \qquad \ln(1- x) \ge - x - x^2,\qquad \forall x \in \left[0,\frac{1}{2} \right].
\end{equation*}
It is also straightforward to verify that
\begin{equation*}
    0 < \frac{2\Delta}{p_0} \le \frac{\frac{1 - \gamma}{12 \gamma}}{\frac{4 \gamma - 1}{3 \gamma }} \le \frac{1}{4} < 1,\qquad \qquad 0 < \frac{2 \Delta}{1 - p_0} \le \frac{\frac{1 - \gamma}{12 \gamma}}{1 - \frac{4 \gamma - 1}{3 \gamma }} \le \frac{1}{8} < \frac{1}{2}.
\end{equation*}
(\ref{eq:lower-derive-e}) comes from the fact that $\gamma \ge \frac{1}{3}$ so $p_0 = \frac{4 \gamma - 1}{3 \gamma} \ge \frac{1}{3}$. (\ref{eq:lower-derive-f}) comes from following:
\begin{equation*}
    \begin{aligned}
        \frac{(1-\gamma)^{-3} (mn-1) \epsilon^{-2}}{mn-1} \cdot \frac{12 \Delta^2}{1-p_0} = \frac{(1-\gamma)^{-3} (mn-1) \epsilon^{-2}}{mn-1} \cdot \frac{12 }{1-\frac{4 \gamma - 1}{3 \gamma }} \cdot \left( \frac{5}{3} \frac{(1-\gamma)^2}{\gamma} \epsilon\right) ^2  = 100 \cdot \frac{1}{\gamma } \le 300.
    \end{aligned}
\end{equation*}
According to the assumption that ALG is $(\epsilon,\delta)$-\textit{smart} on $\mathcal{M}_0$ with the help of prediction $\hat{\boldsymbol{\text{P}}}$, then we have
\begin{equation*}
    \Pr_{\mathcal{M}_0,T} \left( \left\| v^{\hat{\pi}_T} - v^*_{\mathcal{M}_0} \right \|_{\infty}  >\epsilon \right) < \delta.
\end{equation*}
Therefore by (\ref{eq:lower-derive}),
\begin{equation*}
    \Pr_{\mathcal{M}',T} \left( \left\| v^{\hat{\pi}_T} - v^*_{\mathcal{M}'} \right \|_{\infty}  >\epsilon \right) > 2\delta - \Pr_{\mathcal{M}_0,T} \left( \left\| v^{\hat{\pi}_T} - v^*_{\mathcal{M}_0} \right \|_{\infty}  >\epsilon \right) > \delta.
\end{equation*}
This implies that ALG is not $(\epsilon,\delta)$-\textit{smart} on $\mathcal{M}'$. Altogether, the Theorem is proved.

\subsection{More Discussions on the Proof for Theorem \ref{thm:impossibility-result}}

\label{sec:app-discuss-thm-impossibility}

We provide further discussions on the instances $\mathcal{M}_0$, $\mathcal{M}'$. For instance $\mathcal{M}_0$, the state-action pair $(i^{(\text{mi})}_{1,1},a^{\text{(mi)}})$ is the most ``valueable'' pair, meaning that focusing on this pair as frequently as possible leads to the optimal value function. In contrast, for instance $\mathcal{M}'$, the most ``valueable'' pair is $i^{(\text{mi})}_{k',\ell'},a^{\text{(mi)}}$, where $(k',\ell') \ne (1,1)$. Note that the ALG is provided with the same prediction $\hat{\boldsymbol{\text{P}}}$ ($= \boldsymbol{\text{P}}_0$). Thus, while $\hat{\boldsymbol{\text{P}}}$ provides useful information for ALG for learning $\mathcal{M}_0$, it is misleading for learning $\mathcal{M}'$. As a result, no policy can simultaneously confirm that (1) $\hat{\boldsymbol{\text{P}}}$ is useful for learning $\mathcal{M}_0$, and (2) $\hat{\boldsymbol{\text{P}}}$ is misleading for learning $\mathcal{M}'$ and should be ignored entirely. This reasoning ultimately leads to Theorem \ref{thm:impossibility-result}.


\section{Proof in Section 4}

\subsection{Proof for Lemma \ref{lem:property-gradient-v-noupper-unbiase}, \ref{lem:property-gradient-mu-noupper-unbiase}}

\label{sec:app-pf-gradient-vmu-noupper}

We rephrase Lemma \ref{lem:property-gradient-v-noupper-unbiase}, \ref{lem:property-gradient-mu-noupper-unbiase} as Lemma \ref{lem:property-gradient-v-noupper}, Lemma \ref{lem:property-gradient-mu-noupper}, respectively, and provide their proofs.

\begin{lemma}{(Lemma 3 in \cite{jin2020efficiently})}
    The stochastic gradient $\tilde{\boldsymbol{g}}_t^{\boldsymbol{v}}$ for $\boldsymbol{v}$ satisfies $\mathbb{E}[\tilde{\boldsymbol{g}}_t^{\boldsymbol{v}}] = (1-\gamma) \boldsymbol{q} + \boldsymbol{\mu}_t^{\top} (\gamma \boldsymbol{\text{P}} - \hat{\boldsymbol{\text{I}}}) = \boldsymbol{g}^{\boldsymbol{v}}(\boldsymbol{v}_t,\boldsymbol{\mu}_t)$ and $\mathbb{E}[\|\tilde{\boldsymbol{g}}_t^{\boldsymbol{v}}\|_2^2] \le 2$.
    \label{lem:property-gradient-v-noupper}
\end{lemma}

\proof{Proof of Lemma \ref{lem:property-gradient-v-noupper}}
    We can directly compute
    \begin{equation*}
        \mathbb{E}[\tilde{\boldsymbol{g}}_t^{\boldsymbol{v}}] = (1 -\gamma) \boldsymbol{q} + \sum_{i,a,j} \mu_{t,(i,a)} p(j|i,a)(\gamma \boldsymbol{e}_j - \boldsymbol{e_i}) = (1-\gamma) \boldsymbol{q} + \boldsymbol{\mu}^{\top} (\gamma \boldsymbol{\text{P}} - \hat{\boldsymbol{\text{I}}}).
    \end{equation*}
    By definition, it is trivial to check $\|\tilde{\boldsymbol{g}}_t^{\boldsymbol{v}}\|_2^2 \le (1-\gamma)^2 + \gamma^2 + 1 \le 2$ with probability 1.
\endproof

\begin{lemma}
    The stochastic gradient $\tilde{\boldsymbol{g}}_t^{\boldsymbol{\mu}}$ for $\boldsymbol{v}$ satisfies $\mathbb{E}[\tilde{\boldsymbol{g}}_t^{\boldsymbol{\mu}}] = (\hat{\boldsymbol{\text{I}}} - \gamma \boldsymbol{\text{P}}) \boldsymbol{v}_t - \boldsymbol{\text{r}} = \boldsymbol{g}^{\boldsymbol{\mu}}(\boldsymbol{v}_t,\boldsymbol{\mu}_t)$ and $\mathbb{E}[\|\tilde{\boldsymbol{g}}_t^{\boldsymbol{\mu}}-\boldsymbol{g}^{\boldsymbol{\mu}}(\boldsymbol{v},\boldsymbol{\mu})\|_{\infty}^2] \le 9N^2(1-\gamma)^{-2}/t$.
    \label{lem:property-gradient-mu-noupper}
\end{lemma}

\proof{Proof of Lemma \ref{lem:property-gradient-mu-noupper}}
    The proof largely follows the proof for Lemma 4 in \cite{jin2020efficiently}.
    Denote $\tilde{\boldsymbol{g}}_t^{\boldsymbol{\mu,\ell}} = N ( v_{t,i_{\ell}} - \gamma v_{t,j_{\ell}} - \text{r}_{i_{\ell},a_{\ell}}) \boldsymbol{e}_{i_{\ell},a_{\ell}}$, where $(i_{\ell},a_{\ell},j_{\ell})$ be the $\ell-$th pair sampled. Then $\tilde{\boldsymbol{g}}_t^{\boldsymbol{\mu}} = \frac{1}{t}\sum_{\ell=1}^t \tilde{\boldsymbol{g}}_t^{\boldsymbol{\mu,\ell}}$. For each $\tilde{\boldsymbol{g}}_t^{\boldsymbol{\mu,\ell}}$,
    \begin{equation*}
        \mathbb{E}[\tilde{\boldsymbol{g}}_t^{\boldsymbol{\mu,\ell}}] = \sum_{i,a} \sum_j p(j|i,a) (v_i - \gamma v_j - \text{r}_{i,a}) \boldsymbol{e}_{i,a} = (\hat{\boldsymbol{\text{I}}} - \gamma \hat{\boldsymbol{\text{P}}}) \boldsymbol{v} - \boldsymbol{\text{r}} = \boldsymbol{g}^{\boldsymbol{\mu}}(\boldsymbol{v},\boldsymbol{\mu}),
    \end{equation*}
    and $\|\tilde{\boldsymbol{g}}_t^{\boldsymbol{\mu,\ell}}\|_{\infty} \le 3N(1-\gamma)^{-1}$ with probability 1. Hence, $\mathbb{E}[\tilde{\boldsymbol{g}}_t^{\boldsymbol{\mu}}]=\frac{1}{t}\sum_{\ell=1}^t \mathbb{E}[\tilde{\boldsymbol{g}}_t^{\boldsymbol{\mu,\ell}}] =\boldsymbol{g}^{\boldsymbol{\mu}}(\boldsymbol{v},\boldsymbol{\mu}) $, and
    \begin{equation*}
        \mathbb{E}[\|\tilde{\boldsymbol{g}}_t^{\boldsymbol{v}}-\boldsymbol{g}^{\boldsymbol{\mu}}(\boldsymbol{v},\boldsymbol{\mu})\|_{\infty}^2] \le \frac{1}{t} \max_{\ell \in [t]} \mathbb{E}[\|\tilde{\boldsymbol{g}}_t^{\boldsymbol{\mu}}-\boldsymbol{g}^{\boldsymbol{\mu}}(\boldsymbol{v},\boldsymbol{\mu})\|_{\infty}^2] \le \frac{9N^2(1-\gamma)^{-2}}{t}.
    \end{equation*}
\endproof

\subsection{Proof for Lemma \ref{lem:minimax-error-noupper}}

\label{sec:app-pf-minimax-error-noupper}

We derive a stronger version for Lemma \ref{lem:minimax-error-noupper}. We will show that for any $\boldsymbol{v} \in \mathcal{V}$, $\boldsymbol{\mu} \in \mathcal{U}$, 

\begin{equation*}
    \mathbb{E}[f(\bar{\boldsymbol{v}},\boldsymbol{\mu}) - f(\boldsymbol{v},\bar{\boldsymbol{\mu}})] \le 3 \left(\sqrt{|\mathcal{S}|} (1-\gamma)^{-1} + \sqrt{N} \cdot \gamma(1-\gamma)^{-1} \cdot \min \left \{1,\text{Dist} \right\} \right) \cdot \sqrt{\frac{1}{T}} + 9 \sqrt{2} N(1-\gamma)^{-1} \frac{\ln(T)}{T}.
\end{equation*}

This is because
\begin{equation*}
    \mathbb{E}[\text{GAP}(\bar{\boldsymbol{v}},\bar{\boldsymbol{\mu}})] = \mathbb{E}[\max_{\boldsymbol{\mu}'} f(\bar{\boldsymbol{v}},\boldsymbol{\mu}') - \min_{\boldsymbol{v}'} f(\boldsymbol{v}',\bar{\boldsymbol{\mu}})] = \mathbb{E}[f(\bar{\boldsymbol{v}},\bar{\boldsymbol{\mu}}') - f(\bar{\boldsymbol{v}}',\bar{\boldsymbol{\mu}})],
\end{equation*}
where $\bar{\boldsymbol{\mu}}' \in \mathop{\arg\max}_{\boldsymbol{\mu} \in \mathcal{U}} f(\bar{\boldsymbol{v}},\boldsymbol{\mu})$ and $\bar{\boldsymbol{v}}' \in \mathop{\arg\max}_{\boldsymbol{v} \in \mathcal{V}} f(\boldsymbol{v},\bar{\boldsymbol{\mu}})$. Denote $\boldsymbol{g}^{\boldsymbol{v}}(\boldsymbol{v},\boldsymbol{\mu}) = \nabla_{\boldsymbol{v}} f(\boldsymbol{v},\boldsymbol{\mu}) = (1-\gamma) \boldsymbol{q} + \boldsymbol{\mu}^{\top} (\gamma \boldsymbol{\text{P}} - \hat{\boldsymbol{\text{I}}})$,
, $\boldsymbol{g}^{\boldsymbol{\mu}}(\boldsymbol{v},\boldsymbol{\mu}) = - \nabla_{\boldsymbol{\mu}} f(\boldsymbol{v},\boldsymbol{\mu})  = (\hat{\boldsymbol{\text{I}}} - \gamma \boldsymbol{\text{P}}) \boldsymbol{v} - \boldsymbol{\text{r}}$. Denote
\begin{equation*}
    \hat{f}_t^{(\boldsymbol{v})}(\boldsymbol{v},\boldsymbol{\mu}) = (\tilde{\boldsymbol{g}}^{\boldsymbol{v}}_t)^{\top} \boldsymbol{v} + \boldsymbol{\mu}^{\top} \boldsymbol{\text{r}}.
\end{equation*}
We remark that by the construction of $\tilde{\boldsymbol{g}}^{\boldsymbol{v}}$, $\tilde{\boldsymbol{g}}^{\boldsymbol{v}}$ is a function of $\boldsymbol{\mu}$. By Lemma \ref{lem:property-gradient-v-noupper}, we have $\mathbb{E}[\tilde{\boldsymbol{g}}^{\boldsymbol{v}}(\boldsymbol{v},\boldsymbol{\mu})] = (1-\gamma) \boldsymbol{q} + \boldsymbol{\mu}^{\top} (\gamma \boldsymbol{\text{P}} - \hat{\boldsymbol{\text{I}}}) = \boldsymbol{g}^{\boldsymbol{v}}(\boldsymbol{v},\boldsymbol{\mu})$. Similarly, denote
\begin{equation*}
    \hat{f}_t^{(\boldsymbol{\mu})}(\boldsymbol{v},\boldsymbol{\mu}) = (1 - \gamma) \boldsymbol{q}^{\top} \boldsymbol{v} - \boldsymbol{\mu}^{\top} \tilde{\boldsymbol{g}}^{\boldsymbol{\mu}}_t. 
\end{equation*}
$\tilde{\boldsymbol{g}}^{\boldsymbol{\mu}}$ is a function of $\boldsymbol{v}$. By Lemma \ref{lem:property-gradient-mu-noupper}, we have $\mathbb{E}[\tilde{\boldsymbol{g}}^{\boldsymbol{\mu}}(\boldsymbol{v},\boldsymbol{\mu})] = (\hat{\boldsymbol{\text{I}}} - \gamma \boldsymbol{\text{P}}) \boldsymbol{v} - \boldsymbol{\text{r}} = \boldsymbol{g}^{\boldsymbol{\mu}}(\boldsymbol{v},\boldsymbol{\mu})$.

Now, we decompose $\mathbb{E}[f(\bar{\boldsymbol{v}},\boldsymbol{\mu}) - f(\boldsymbol{v},\bar{\boldsymbol{\mu}})]$ as follows: For any $\boldsymbol{v} \in \mathcal{V}$, $\boldsymbol{\mu} \in \mathcal{U}$,
\begin{subequations}
    \begin{align}
        & \mathbb{E} \left[ f(\bar{\boldsymbol{v}},\boldsymbol{\mu}) - f(\boldsymbol{v},\bar{\boldsymbol{\mu}})\right] \nonumber \\
        = & \mathbb{E} \left[\frac{1}{T} \sum_{t=1}^T f(\boldsymbol{v}_t,\boldsymbol{\mu}) - \frac{1}{T} \sum_{t=1}^T f(\boldsymbol{v},\boldsymbol{\mu}_t)\right] \label{eq:proof-minimax-error-noupper-decom-a} \\
        = & \frac{1}{T} \mathbb{E}\left [ \sum_{t=1}^T  f(\boldsymbol{v}_t,\boldsymbol{\mu}) -  \sum_{t=1}^T f(\boldsymbol{v}_t,\boldsymbol{\mu}_t)  \right]+ \frac{1}{T}\mathbb{E}\left [ \sum_{t=1}^T f(\boldsymbol{v}_t,\boldsymbol{\mu}_t)- \sum_{t=1}^T f(\boldsymbol{v},\boldsymbol{\mu}_t) \right] \nonumber \\
        \le & \frac{1}{T} \mathbb{E}\left [ \sum_{t=1}^T  \hat{f}^{(\boldsymbol{\mu})}_t(\boldsymbol{v}_t,\boldsymbol{\mu}) -  \sum_{t=1}^T \hat{f}^{(\boldsymbol{\mu})}_t(\boldsymbol{v}_t,\boldsymbol{\mu}_t)  \right] + \frac{1}{T}\mathbb{E}\left [ \sum_{t=1}^T \hat{f}^{(\boldsymbol{v})}_t(\boldsymbol{v}_t,\boldsymbol{\mu}_t)- \sum_{t=1}^T \hat{f}^{(\boldsymbol{v})}_t(\boldsymbol{v},\boldsymbol{\mu}_t) \right].\label{eq:proof-minimax-error-noupper-decom-b}
    \end{align}
    \label{eq:proof-minimax-error-noupper-decompose}
\end{subequations}
(\ref{eq:proof-minimax-error-noupper-decom-a}) comes from the bilinear structure of $f(\cdot,\cdot)$. The first term in (\ref{eq:proof-minimax-error-noupper-decom-b}) comes from the following:
\begin{equation*}
    \begin{aligned}
        & \mathbb{E}\left [ \sum_{t=1}^T  f(\boldsymbol{v}_t,\boldsymbol{\mu}) -  \sum_{t=1}^T f(\boldsymbol{v}_t,\boldsymbol{\mu}_t)  \right] \\
        =& \mathbb{E}\left [ \sum_{t=1}^T  f(\boldsymbol{v}_t,\boldsymbol{\mu}) - \sum_{t=1}^T  \hat{f}^{(\boldsymbol{\mu})}_t(\boldsymbol{v}_t,\boldsymbol{\mu})+ \sum_{t=1}^T  \hat{f}^{(\boldsymbol{\mu})}_t(\boldsymbol{v}_t,\boldsymbol{\mu})- \sum_{t=1}^T \hat{f}^{(\boldsymbol{\mu})}_t(\boldsymbol{v}_t,\boldsymbol{\mu}_t) +\sum_{t=1}^T \hat{f}^{(\boldsymbol{\mu})}_t(\boldsymbol{v}_t,\boldsymbol{\mu}_t) - \sum_{t=1}^T f(\boldsymbol{v}_t,\boldsymbol{\mu}_t)  \right] \\
        =& \mathbb{E}\left [ \sum_{t=1}^T  \hat{f}^{(\boldsymbol{\mu})}_t(\boldsymbol{v}_t,\boldsymbol{\mu}) -  \sum_{t=1}^T \hat{f}^{(\boldsymbol{\mu})}_t(\boldsymbol{v}_t,\boldsymbol{\mu}_t)  \right] + \mathbb{E}\left[ \sum_{t=1}^T \left(\tilde{\boldsymbol{g}}^{\boldsymbol{\mu}}_t-\boldsymbol{g}^{\boldsymbol{\mu}}(\boldsymbol{v}_t,\boldsymbol{\mu})  \right)^{\top} \boldsymbol{\mu} \right] + \mathbb{E}\left[ \sum_{t=1}^T \left(\boldsymbol{g}^{\boldsymbol{\mu}}(\boldsymbol{v}_t,\boldsymbol{\mu}_t) - \tilde{\boldsymbol{g}}^{\boldsymbol{\mu}}_t \right)^{\top} \boldsymbol{\mu}_t \right] \\
         = &\mathbb{E}\left [ \sum_{t=1}^T  \hat{f}^{(\boldsymbol{\mu})}_t(\boldsymbol{v}_t,\boldsymbol{\mu}) -  \sum_{t=1}^T \hat{f}^{(\boldsymbol{\mu})}_t(\boldsymbol{v}_t,\boldsymbol{\mu}_t)  \right].
    \end{aligned}
\end{equation*}
The last inequality comes from fact that $\boldsymbol{g}^{\boldsymbol{\mu}}(\boldsymbol{v},\boldsymbol{\mu}) = (\hat{\boldsymbol{\text{I}}} - \gamma \hat{\boldsymbol{\text{P}}}) \boldsymbol{v} - \boldsymbol{\text{r}}$ does not depend on $\boldsymbol{\mu}$, and the conditional expectation that $\forall t$, $\boldsymbol{\mu}' \in \mathcal{U}$, 
\begin{equation*}
\mathbb{E} \left[\left(\tilde{\boldsymbol{g}}^{\boldsymbol{\mu}}_t-\boldsymbol{g}^{\boldsymbol{\mu}}(\boldsymbol{v}_t,\boldsymbol{\mu})  \right)^{\top} \boldsymbol{\mu}' \right] = \mathbb{E} \left[\mathbb{E} \left[ \left(\tilde{\boldsymbol{g}}^{\boldsymbol{\mu}}_t-\boldsymbol{g}^{\boldsymbol{\mu}}(\boldsymbol{v}_t,\boldsymbol{\mu})  \right)^{\top} \boldsymbol{\mu}' | \sigma(\{(i_k,a_k,j_k)\}_{k=1}^{t-1})\right] \right] = 0.
\end{equation*}
Similarly, the second term in (\ref{eq:proof-minimax-error-noupper-decom-b}) comes from the following:
\begin{equation*}
    \begin{aligned}
    & \mathbb{E}\left [ \sum_{t=1}^T f(\boldsymbol{v}_t,\boldsymbol{\mu}_t)- \sum_{t=1}^T f(\boldsymbol{v},\boldsymbol{\mu}_t) \right] \\
    = & \mathbb{E}\left [ \sum_{t=1}^T f(\boldsymbol{v}_t,\boldsymbol{\mu}_t) - \sum_{t=1}^T \hat{f}^{(\boldsymbol{v})}_t(\boldsymbol{v}_t,\boldsymbol{\mu}_t) + \sum_{t=1}^T \hat{f}^{(\boldsymbol{v})}_t(\boldsymbol{v}_t,\boldsymbol{\mu}_t)- \sum_{t=1}^T \hat{f}^{(\boldsymbol{v})}_t(\boldsymbol{v},\boldsymbol{\mu}_t) + \sum_{t=1}^T \hat{f}^{(\boldsymbol{v})}_t(\boldsymbol{v},\boldsymbol{\mu}_t)- \sum_{t=1}^T 
 f(\boldsymbol{v},\boldsymbol{\mu}_t) \right] \\
    = & \mathbb{E} \left[ \sum_{t=1}^T \hat{f}^{(\boldsymbol{v})}_t(\boldsymbol{v}_t,\boldsymbol{\mu}_t)- \sum_{t=1}^T \hat{f}^{(\boldsymbol{v})}_t(\boldsymbol{v},\boldsymbol{\mu}_t)\right] + \mathbb{E}\left[ \sum_{t=1}^T \left(\boldsymbol{g}^{\boldsymbol{v}}(\boldsymbol{v}_t,\boldsymbol{\mu}_t) - \tilde{\boldsymbol{g}}^{\boldsymbol{v}}_t \right)^{\top} \boldsymbol{v}_t \right] + \mathbb{E}\left[ \sum_{t=1}^T \left(\tilde{\boldsymbol{g}}^{\boldsymbol{v}}_t - \boldsymbol{g}^{\boldsymbol{v}}(\boldsymbol{v},\boldsymbol{\mu}_t) \right)^{\top} \boldsymbol{v} \right] \\
    = &\mathbb{E} \left[ \sum_{t=1}^T \hat{f}^{(\boldsymbol{v})}_t(\boldsymbol{v}_t,\boldsymbol{\mu}_t)- \sum_{t=1}^T \hat{f}^{(\boldsymbol{v})}_t(\boldsymbol{v},\boldsymbol{\mu}_t)\right].
    \end{aligned}
\end{equation*}

Now we analyze the two terms of (\ref{eq:proof-minimax-error-noupper-decom-b}), respectively.

\textbf{Analysis for $\mathbb{E}\left [ \sum_{t=1}^T  f(\boldsymbol{v}_t,\boldsymbol{\mu}) -  \sum_{t=1}^T f(\boldsymbol{v}_t,\boldsymbol{\mu}_t)  \right]$:} Notice that the update fule for $\boldsymbol{\mu}_t$ in (\ref{eq:smd-prediction-noupper-mu-update}) is equivalent to the following optimistic online mirror descent update rule (Algorithm 6.4 in \cite{orabona2019modern}): 
\begin{equation*}
    \boldsymbol{\mu}_{t+1} = \mathop{\arg\min}_{\boldsymbol{\mu} \in \mathcal{U}} \left \{ \left(\tilde{\boldsymbol{g}}_t^{\boldsymbol{\mu}}  - \bar{\boldsymbol{g}}_t^{\boldsymbol{\mu}} + \bar{\boldsymbol{g}}_{t+1}^{\boldsymbol{\mu}} \right)^{\top} \boldsymbol{\mu} + \frac{1}{\eta_t^{\mu}} B_\psi(\boldsymbol{\mu};\boldsymbol{\mu}_t)  \right \}.
\end{equation*}
where $B_{\psi}(\boldsymbol{z};\boldsymbol{y}) = \sum_{i \in [N]} z_i \ln \frac{z_i}{y_i}$. Then we have
\begin{subequations}
    \begin{align}
    \sum_{t=1}^T  \hat{f}^{(\boldsymbol{\mu})}_t(\boldsymbol{v}_t,\boldsymbol{\mu}) -  \sum_{t=1}^T \hat{f}^{(\boldsymbol{\mu})}_t(\boldsymbol{v}_t,\boldsymbol{\mu}_t) & = \sum_{t=1}^T (\tilde{\boldsymbol{g}}^{\boldsymbol{\mu}}(\boldsymbol{v}_t,\boldsymbol{\mu}_t))^{\top} \boldsymbol{\mu}_t - \sum_{t=1}^T (\tilde{\boldsymbol{g}}^{\boldsymbol{\mu}}(\boldsymbol{v},\boldsymbol{\mu}_t))^{\top} \boldsymbol{\mu} \nonumber \\
    & = \sum_{t=1}^T (\tilde{\boldsymbol{g}}^{\boldsymbol{\mu}}(\boldsymbol{v}_t,\boldsymbol{\mu}_t))^{\top} \boldsymbol{\mu}_t - \sum_{t=1}^T (\tilde{\boldsymbol{g}}^{\boldsymbol{\mu}}(\boldsymbol{v}_t,\boldsymbol{\mu}_t))^{\top} \boldsymbol{\mu} \label{eq:proof-minimax-error-noupper-mu-oco-regret-a}\\
    & \le \frac{\ln(N)}{\eta_T^{\mu}} + \frac{1}{2} \sum_{t=1}^T \eta_t^{\mu} \|\tilde{\boldsymbol{g}}_t^{\boldsymbol{\mu}}  - \bar{\boldsymbol{g}}_t^{\boldsymbol{\mu}}\|_{\infty}^2 \label{eq:proof-minimax-error-noupper-mu-oco-regret-b}\\
    & = \sqrt{2} \cdot \sqrt{\ln(N)} \cdot \sqrt{\sum_{t=1}^T \|\tilde{\boldsymbol{g}}_t^{\boldsymbol{\mu}}  - \bar{\boldsymbol{g}}_t^{\boldsymbol{\mu}}\|_{\infty}^2} \nonumber \\
    & + \frac{\sqrt{2}}{4} \cdot \sqrt{\ln(N)} \cdot \sum_{t=1}^T \frac{\|\tilde{\boldsymbol{g}}_t^{\boldsymbol{\mu}}  - \bar{\boldsymbol{g}}_t^{\boldsymbol{\mu}}\|_{\infty}^2}{\sqrt{\sum_{i=1}^t \|\tilde{\boldsymbol{g}}^{\boldsymbol{\mu}}_i - \bar{\boldsymbol{g}}^{\boldsymbol{\mu}}_i\|_{\infty}^2}} \label{eq:proof-minimax-error-noupper-mu-oco-regret-c}\\
    & \le \frac{3}{2}\sqrt{2} \cdot \sqrt{\ln(N)} \cdot \sqrt{\sum_{t=1}^T \|\tilde{\boldsymbol{g}}_t^{\boldsymbol{\mu}}  - \bar{\boldsymbol{g}}_t^{\boldsymbol{\mu}}\|_{\infty}^2}. \label{eq:proof-minimax-error-noupper-mu-oco-regret-d}
    \end{align} 
    \label{eq:proof-minimax-error-noupper-mu-oco-regret}
\end{subequations}
(\ref{eq:proof-minimax-error-noupper-mu-oco-regret-a}) comes from the fact that $\tilde{\boldsymbol{g}}^{\boldsymbol{\mu}}$ only depends on $\boldsymbol{v}$ based on its construction rule. (\ref{eq:proof-minimax-error-noupper-mu-oco-regret-b}) comes from Theorem 6.20 in \cite{orabona2019modern} (we take $\ell_t(\boldsymbol{x}) = \tilde{\boldsymbol{g}}_t^{\boldsymbol{\mu},\top}\boldsymbol{x}$, and $\bar{\boldsymbol{g}}_t$ as the prediction of the subgradient). (\ref{eq:proof-minimax-error-noupper-mu-oco-regret-c}) comes from the definition of $\eta_t^{\mu}$. (\ref{eq:proof-minimax-error-noupper-mu-oco-regret-d}) comes from Lemma 4.13 in \cite{orabona2019modern}. Therefore,

\begin{subequations}
    \begin{align}
    & \mathbb{E}\left [ \sum_{t=1}^T  \hat{f}^{(\boldsymbol{\mu})}_t(\boldsymbol{v}_t,\boldsymbol{\mu}) -  \sum_{t=1}^T \hat{f}^{(\boldsymbol{\mu})}_t(\boldsymbol{v}_t,\boldsymbol{\mu}_t)  \right]  \nonumber \\
    \le & \frac{3}{2}\sqrt{2} \cdot \sqrt{\ln(N)} \cdot \sqrt{\mathbb{E} \left[\sum_{t=1}^T \|\tilde{\boldsymbol{g}}_t^{\boldsymbol{\mu}}  - \bar{\boldsymbol{g}}_t^{\boldsymbol{\mu}}\|_{\infty}^2\right]} \label{eq:proof-minimax-error-noupper-mu-error-a} \\
     \le & 3 \sqrt{\ln(N)}  \sqrt{\mathbb{E} \left[\sum_{t=1}^T  \|\tilde{\boldsymbol{g}}_t^{\boldsymbol{\mu}}  - \boldsymbol{g}^{\boldsymbol{\mu}}(\boldsymbol{v}_t,\boldsymbol{\mu}_t) \|_{\infty}^2\right] + \mathbb{E} \left[\sum_{t=1}^T  \| \boldsymbol{g}^{\boldsymbol{\mu}}(\boldsymbol{v}_t,\boldsymbol{\mu}_t) -\bar{\boldsymbol{g}}_t^{\boldsymbol{\mu}}\|_{\infty}^2\right]} \label{eq:proof-minimax-error-noupper-mu-error-b}\\
    = & 3 \sqrt{\ln(N)}  \sqrt{\sum_{t=1}^T  \mathbb{E}\left[\|\tilde{\boldsymbol{g}}_t^{\boldsymbol{\mu}}  - \boldsymbol{g}^{\boldsymbol{\mu}}(\boldsymbol{v}_t,\boldsymbol{\mu}_t) \|_{\infty}^2\right ] + \sum_{t=1}^T  \mathbb{E} \left [\| \boldsymbol{g}^{\boldsymbol{\mu}}(\boldsymbol{v}_t,\boldsymbol{\mu}_t) -\bar{\boldsymbol{g}}_t^{\boldsymbol{\mu}}\|_{\infty}^2\right]} \nonumber \\
    \le & 3 \sqrt{\ln(N)}  \sqrt{\sum_{t=1}^T  \frac{9N^2(1-\gamma)^{-2}}{t} + T \cdot N \cdot \gamma^2(1-\gamma)^{-2} \cdot \min \left \{1,\text{Dist}^2 \right\}}  \label{eq:proof-minimax-error-noupper-mu-error-c} \\
    \le & 3 \sqrt{\ln(N)}  \sqrt{18N^2(1-\gamma)^{-2}\cdot 
    \ln(T) + T \cdot N \cdot \gamma^2(1-\gamma)^{-2} \cdot \min \left \{1,\text{Dist}^2 \right\}} \nonumber
    \end{align}
    \label{eq:proof-minimax-error-noupper-mu-error}
\end{subequations}
(\ref{eq:proof-minimax-error-noupper-mu-error-a}) comes from Jensen Inequality. (\ref{eq:proof-minimax-error-noupper-mu-error-b}) comes from the fact that for any two vectors $\boldsymbol{a}$, $\boldsymbol{b}$, $\|\boldsymbol{a}+\boldsymbol{b}\|_{\infty}^2 \le 2 \|\boldsymbol{a}\|_{\infty}^2 + 2 \|\boldsymbol{b}\|_{\infty}^2$. (\ref{eq:proof-minimax-error-noupper-mu-error-c}) comes from Lemma \ref{lem:property-gradient-mu-noupper} and the following calculation:
\begin{equation*}
\boldsymbol{g}^{\boldsymbol{\mu}}(\boldsymbol{v}_t,\boldsymbol{\mu}_t) -\bar{\boldsymbol{g}}_t^{\boldsymbol{\mu}} = (\hat{\boldsymbol{\text{I}}} - \gamma \boldsymbol{\text{P}}) \boldsymbol{v}_{t} - \boldsymbol{\text{r}} - \left((\hat{\boldsymbol{\text{I}}} - \gamma \hat{\boldsymbol{\text{P}}}) \boldsymbol{v}_{t} - \boldsymbol{\text{r}} \right) = \gamma (\boldsymbol{\text{P}} - \hat{\boldsymbol{\text{P}}})\boldsymbol{v}_t,
\end{equation*}
so
\begin{equation*}
    \mathbb{E} \left [\| \boldsymbol{g}^{\boldsymbol{\mu}}(\boldsymbol{v}_t,\boldsymbol{\mu}_t) -\bar{\boldsymbol{g}}_t^{\boldsymbol{\mu}}\|_{\infty}^2\right] \le N \cdot \gamma^2(1-\gamma)^{-2} \cdot \min \left \{1,\text{Dist}^2 \right\}.
\end{equation*}

\textbf{Analysis for $\mathbb{E}\left [ \sum_{t=1}^T f(\boldsymbol{v}_t,\boldsymbol{\mu}_t)- \sum_{t=1}^T f(\boldsymbol{v},\boldsymbol{\mu}_t) \right]$:} This analysis largely follows the line as the previous part. Notice that the update rule for $\boldsymbol{v}_t$ in (\ref{eq:smd-prediction-noupper-v-update}) is equivalent to the following online mirror descent (OMD) rule:
\begin{equation*}
    \boldsymbol{v}_{t+1}  = \mathop{\arg\min}_{\boldsymbol{v} \in \mathcal{V}} \left \{ (\tilde{\boldsymbol{g}}^{\boldsymbol{v}}_t)^{\top} \boldsymbol{v} + \frac{1}{2 \eta_t} \|\boldsymbol{v}_t - \boldsymbol{v}\|_2^2 \right \}.
\end{equation*}
Then we have
\begin{align*}
    \sum_{t=1}^T \hat{f}^{(\boldsymbol{v})}(\boldsymbol{v}_t,\boldsymbol{\mu}_t)- \sum_{t=1}^T \hat{f}^{(\boldsymbol{v})}(\boldsymbol{v},\boldsymbol{\mu}_t) & = \sum_{t=1}^T (\tilde{\boldsymbol{g}}^{\boldsymbol{v}}(\boldsymbol{v}_t,\boldsymbol{\mu}_t))^{\top} \boldsymbol{v}_t - \sum_{t=1}^T (\tilde{\boldsymbol{g}}^{\boldsymbol{v}}(\boldsymbol{v},\boldsymbol{\mu}_t))^{\top} \boldsymbol{v} \\
    & = \sum_{t=1}^T (\tilde{\boldsymbol{g}}^{\boldsymbol{v}}(\boldsymbol{v}_t,\boldsymbol{\mu}_t))^{\top} \boldsymbol{v}_t - \sum_{t=1}^T (\tilde{\boldsymbol{g}}^{\boldsymbol{v}}(\boldsymbol{v}_t,\boldsymbol{\mu}_t))^{\top} \boldsymbol{v} \\
    & \le \frac{|\mathcal{S}|(1-\gamma)^{-2}}{\eta_T^v} + \frac{1}{2} \sum_{t=1}^T \eta_t^v \|\tilde{\boldsymbol{g}}^{\boldsymbol{v}}(\boldsymbol{v}_t,\boldsymbol{\mu}_t)\|_2^2 \\
    & = \sqrt{2} \cdot \sqrt{|\mathcal{S}|} (1-\gamma)^{-1} \cdot \sqrt{\sum_{t=1}^T\|\tilde{\boldsymbol{g}}^{\boldsymbol{v}}(\boldsymbol{v}_t,\boldsymbol{\mu}_t)\|_2^2} \\
    & + \frac{\sqrt{2}}{4} \cdot  \sqrt{|\mathcal{S}|} (1-\gamma)^{-1} \cdot \sum_{t=1}^T \frac{\|\tilde{\boldsymbol{g}}^{\boldsymbol{v}}(\boldsymbol{v}_t,\boldsymbol{\mu}_t)\|_2^2}{\sqrt{\sum_{i=1}^t \|\tilde{\boldsymbol{g}}^{\boldsymbol{v}}(\boldsymbol{v}_i,\boldsymbol{\mu}_i)\|_2^2}} \\
    & \le \frac{3}{2} \sqrt{2} \cdot \sqrt{|\mathcal{S}|} (1-\gamma)^{-1} \cdot \sqrt{\sum_{t=1}^T\|\tilde{\boldsymbol{g}}^{\boldsymbol{v}}(\boldsymbol{v}_t,\boldsymbol{\mu}_t)\|_2^2},
\end{align*}
where the second equality comes from the fact that $\tilde{\boldsymbol{g}}^{\boldsymbol{v}}$ only depends on $\boldsymbol{\mu}$ based on its construction rule. The first inequality comes from Theorem 6.10 in \cite{orabona2019modern}. The third equality comes from the definition of $\eta_t^{v}$. The second inequality comes from Lemma 4.13 in \cite{orabona2019modern}. Therefore,
\begin{equation}
\begin{aligned}
\mathbb{E} \left[ \sum_{t=1}^T \hat{f}^{(\boldsymbol{v})}(\boldsymbol{v}_t,\boldsymbol{\mu}_t)- \sum_{t=1}^T \hat{f}^{(\boldsymbol{v})}(\boldsymbol{v},\boldsymbol{\mu}_t)\right] & \le \frac{3}{2} \sqrt{2} \cdot \sqrt{|\mathcal{S}|} (1-\gamma)^{-1} \cdot    \sqrt{\sum_{t=1}^T \mathbb{E}\left[\|\tilde{\boldsymbol{g}}^{\boldsymbol{v}}(\boldsymbol{v}_t,\boldsymbol{\mu}_t)\|_2^2\right]} \\
& \le 3 \sqrt{|\mathcal{S}|} (1-\gamma)^{-1} \sqrt{T},
\end{aligned}
\label{eq:proof-minimax-error-noupper-v-error}
\end{equation}
where the first inequality comes from Jensen Inequality. The second inequality comes from Lemma \ref{lem:property-gradient-v-noupper}. Consequently, combine (\ref{eq:proof-minimax-error-noupper-v-error}), (\ref{eq:proof-minimax-error-noupper-mu-error}) into (\ref{eq:proof-minimax-error-noupper-decompose}), we finish the proof.

\subsection{Proof for Theorem \ref{thm:sample-complexity-noupper}}

\label{sec:app-pf-sample-complexity}

The proof of theorem requires the following Lemma from \cite{jin2020efficiently}, which states that an expected $\epsilon$-optimal solution for the minimax problem (\ref{eq:lp-minimax-bellman-dmdp}) implies an expected $O((1-\gamma)^{-1}\epsilon)$-optimal policy:

\begin{lemma}{(Lemma 9, \cite{jin2020efficiently})}
    Given an expected $\epsilon$-optimal solution for minimax problem (\ref{eq:lp-minimax-bellman-dmdp}) with initial distribution $\boldsymbol{q}$, dubbed as $(\boldsymbol{v}^{(\epsilon)},\boldsymbol{\mu}^{(\epsilon)})$, let $\pi^{(\epsilon)}$ satisfies $\mu_{(i,a)}^{(\epsilon)} = \lambda_i \pi_{(i,a)}^{(\epsilon)}$ for some $\boldsymbol{\lambda} \in \Delta^{|\mathcal{S}|}$ and $\pi_i^{(\epsilon)} \in \Delta^{|\mathcal{A}_i|}$, $\forall i \in \mathcal{S}$.
    Then $ v^*(\boldsymbol{q}) \le \mathbb{E}[v^{\boldsymbol{\pi}^{(\epsilon)}}(\boldsymbol{q})]  + 3  (1-\gamma)^{-1} \epsilon$.
    \label{lem:solution-to-policy}
\end{lemma}

By Lemma \ref{lem:solution-to-policy}, an expected $(1-\gamma) \epsilon/3$-optimal solution for minimax problem (\ref{eq:lp-minimax-bellman-dmdp}) implies an expected $\epsilon$-optimal policy. Choose an appropriate optimization length $T$ such that $\text{Err}_{v}$, $\text{Err}_{\mu,1}$, $\text{Err}_{\mu,2}$ satisfy $\le (1-\gamma) \epsilon/9$. Let the required solution be $T_{v}',T_{\mu,1}',T_{\mu,2}'$, respectively. For any optimization length $T \ge \max \{T_{v}',T_{\mu,1}',T_{\mu,2}' \}$, the algorithm outputs an expected $\epsilon$-optimal policy. Additionally, each step in the optimization process requires exactly 2 samples of state transitions (construct $\tilde{\boldsymbol{g}}^{\boldsymbol{v}}_t, \tilde{\boldsymbol{g}}^{\boldsymbol{\mu}}_t$). Altogether, the Theorem is established.

\section{Details for Numerical Experiments}

\label{sec:app-details-numerical}
\subsection{MDP instance}

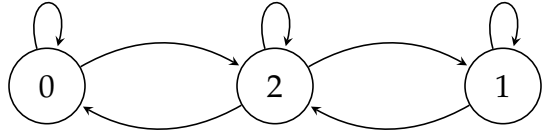
\begin{figure}[htb]
\centering
\begin{tikzpicture}[->,>=stealth,shorten >=1pt,auto,node distance=2cm,semithick]
    \node[circle, draw, minimum size=1cm] (A) at (0,0) {0};
    \node[circle, draw, minimum size=1cm] (B) at (3,0) {2};
    \node[circle, draw, minimum size=1cm] (C) at (6,0) {1};

    \path[black] (A) edge[loop above] (A);
    \path[black] (B) edge[loop above] (B);
    \path[black] (C) edge[loop above] (C);

    \path[black] (A) edge[bend left] (B);
    \path[black] (B) edge[bend left] (C);
    \path[black] (C) edge[bend left] (B);
    \path[black] (B) edge[bend left] (A);
\end{tikzpicture}
\caption{MDP Example}
\label{fig:mdp-example}
\end{figure}

We consider the following MDP instance, which is illustrated in Figure \ref{fig:mdp-example}. For state 0, it can choose two actions: ``stay'', ``leave''. If it chooses ``stay'', then it receives reward 0.001 and stays in state 0 with probability 1. If it chooses ``leave'', it receives reward 0.5. Then, it stays in state 0 with probability 0.4 and transits to state 2 with probability 0.6. Similarly, for state 1, it can choose two actions: ``stay'', ``leave''. If it chooses ``stay'', then it receives reward 0.001 and stays in state 1 with probability 1. If it chooses ``leave'', it receives reward 0.5. Then, it stays in state 1 with probability 0.4 and transits to state 2 with probability 0.6. For state 2, it can choose two actions: ``left'', ``right''. If it chooses ``left'', then it receives reward 1 with probability 1, stays in state 2 with probability 0.2, or transits to state 0, state 1 with probability 0.4, respectively. If it chooses ``right'', then it receives reward 1 with probability 1, stays in state 2 with probability 0.6, or transits to state 0, state 1 with probability 0.2, respectively. Then, the space of state $\mathcal{S} = \{0,1,2\}$, state-action pair 
$$\mathcal{N} = \{(0,\text{stay}),(0,\text{leave}),(1,\text{stay}),(1,\text{leave}),(2,\text{left}),(2,\text{right})\}.$$ The reward vector $\boldsymbol{\text{r}} = (0.001,0.5,0.001,0.5,1,1)$, and transition matrix is
\begin{equation*}
    \boldsymbol{\text{P}} = \begin{pmatrix}
        1 & 0 & 0 \\
        0.4 & 0 & 0.6 \\
        0 & 1 & 0 \\
        0 & 0.4 & 0.6\\
        0.4 & 0.4 & 0.2 \\
        0.2 & 0.2 & 0.6
    \end{pmatrix}.
\end{equation*}

It is straightforward to see that the optimal policy for each state is ``leave'' for state 0 and 1, and ``right'' for state 2, since the action that possibly results in arriving in state 2 and staying in state 2 will always bring in higher reward.

\subsection{Experimental Settings}

We set $\gamma = 0.5$, $\epsilon = 0.05$, $\boldsymbol{q} = (0.4,0.4,0.2)$. We vary the number of samples $T$ from 100 to 16000. When applying OpPMD-NAC, we set the prediction matrix as
\begin{equation*}
    \hat{\boldsymbol{\text{P}}} = \begin{pmatrix}
        0 & 1 & 0 \\
        0 & 1 & 0 \\
        1 & 0 & 0 \\
        1 & 0 & 0\\
        1 & 0 & 0 \\
        0 & 1 & 0
    \end{pmatrix}.
\end{equation*}
It is straightforward to verify $\text{Dist}(\boldsymbol{\text{P}},\hat{\boldsymbol{\text{P}}}) > 1$.

\section{Discussions on Future Directions}

\label{sec:app-discussions-future}

There are several interesting future directions.

\textbf{Extensions to other MDP or RL models} It is interesting to investigate other MDP or RL models with predictions on the transition matrix, such as average-reward MDPs (AMDPs), online tabular finite-horizon episodic MDPs and constrained MDPs. 

\textbf{Other Forms of Predictions} Another intriguing direction is exploring whether other forms of predictions can enhance the process of solving MDPs. For example, \cite{golowich2022can} examines online tabular finite-horizon episodic MDP with prediction on Q-value, and \cite{li2024beyond} considers single-trajectory time-varying MDPs with machine-learned prediction. Additionally, since such advice or predictions often originate from historical observations of other MDP models, such as robotics, it is interesting to see whether these possibly biased observations or datasets can be directly to use to enhance algorithm performance and improve sample complexity bound in solving MDPs.

\end{document}